%% file: main.tex
\documentclass[conference,9pt]{IEEEtran}

\usepackage{algpseudocode}
\algtext*{EndWhile}
\algtext*{EndIf}
\algtext*{EndFor}
\algtext*{EndProcedure}
\usepackage{amsmath}
\usepackage{amssymb}
\usepackage{array}
\usepackage{algorithm}
\usepackage{bm}
\usepackage[bookmarks=false, hidelinks]{hyperref}
\usepackage[noadjust]{cite}
\usepackage[font=small]{caption}
\usepackage{color}
\usepackage{comment}
\usepackage{dblfloatfix}
\usepackage[inline]{enumitem}
\usepackage{epsfig}
\usepackage{etoolbox}
\usepackage{fancybox}
\usepackage{float}
\usepackage{fixltx2e}
\usepackage{graphicx}
\usepackage{listings}
\usepackage{mathrsfs}
\usepackage{multirow}
\usepackage{multicol}
\usepackage{pifont}
\usepackage{placeins}
\usepackage{rotating}
\usepackage{setspace}
\usepackage{soul}
\usepackage[hang, tight]{subfigure}
\usepackage{threeparttable}
\usepackage{times}
\usepackage{url}
\usepackage{upgreek}
\usepackage{verbatim}
\usepackage{wrapfig}
\usepackage[usenames,dvipsnames]{xcolor}
\usepackage{authblk}
\usepackage[]{siunitx}
\usepackage[super]{nth}
\usepackage{tabularx}
\usepackage{supertabular}
\usepackage{tikz}
\usepackage{bbm}
\usepackage{blindtext}

\graphicspath{{./_fig/}}

\usepackage{bbding}

\makeatletter
\renewcommand\footnoterule{%
  \kern-3\p@
  \hrule\@width0.4\columnwidth
  \kern2.6\p@}
  \makeatother

\usepackage[bottom,hang,flushmargin]{footmisc}

\usepackage{tikz}
\usepackage{xcolor}
\newcommand*\circled[1]{\tikz[baseline=(char.base)]{
            \node[shape=circle,fill,inner sep=1pt,scale=0.8] (char) {\textcolor{white}{#1}};}}

\begin{document}





\title{\vspace{-.18in}PANDA: Architecture-Level \underline{P}ower Evaluation by Unifying \underline{A}nalytical a\underline{nd} M\underline{a}chine Learning Solutions\vspace{-.11in}}

\author[]{ \fontsize{11}{11}\selectfont Qijun Zhang$^1$, Shiyu Li$^2$, Guanglei Zhou$^2$, Jingyu Pan$^2$, Chen-Chia Chang$^2$, Yiran Chen$^2$, Zhiyao Xie$^1$\textsuperscript{*}\vspace{-6pt}}

\affil[]{\fontsize{10}{10}\selectfont $^1$Hong Kong University of Science and Technology, $^2$Duke University\vspace{-7pt}}

\affil[]{	qzhangcs@connect.ust.hk,
         \{shiyu.li, guanglei.zhou, jingyu.pan, chenchia.chang, yiran.chen\}@duke.edu, eezhiyao@ust.hk\vspace{-12pt}}

\maketitle
\begingroup\renewcommand\thefootnote{*}
\footnotetext{Corresponding Author}
\endgroup

\input{_txt/abstract}

\input{_txt/1_introduction}


\input{_txt/4_Methodology}

\input{_txt/5_experiment}

\input{_txt/6_discussion}

\input{_txt/6_conclusion}

\bibliographystyle{IEEEtran}
\bibliography{references_1, references_2, references_3}
\end{document}

%% file: _txt/abstract.tex
\begin{abstract}

Power efficiency is a critical design objective in modern microprocessor design. To evaluate the impact of architectural-level design decisions, an accurate yet efficient architecture-level power model is desired. However, widely adopted data-independent analytical power models like McPAT and Wattch have been criticized for their unreliable accuracy. While some machine learning (ML) methods have been proposed for architecture-level power modeling, they rely on sufficient known designs for training and perform poorly when the number of available designs is limited, which is typically the case in realistic scenarios.

In this work, we derive a general formulation that unifies existing architecture-level power models. Based on the formulation, we propose PANDA, an innovative architecture-level solution that combines the advantages of analytical and ML power models. It achieves unprecedented high accuracy on unknown new designs even when there are very limited designs for training, which is a common challenge in practice. Besides being an excellent power model, it can predict area, performance, and energy accurately. PANDA further supports power prediction for unknown new technology nodes. In our experiments, besides validating the superior performance and the wide range of functionalities of PANDA, we also propose an application scenario, where PANDA proves to identify high-performance design configurations given a power constraint.

\end{abstract}

%% file: _txt/1_introduction.tex
\section{Introduction}





Power efficiency is a critical design objective in modern microprocessor design. With the continuous growth in chip complexity, optimizing designs for better power efficiency requires a significant amount of manpower and long turnaround time. Therefore, there is a high demand for fast, yet high-fidelity early-stage power modeling techniques to facilitate efficient design optimizations. For instance, chip architects may need to efficiently evaluate the power efficiency, as well as other design qualities of multiple new design configurations at the architecture-level, before starting the detailed register-transfer level (RTL) implementation and subsequent VLSI design flow.


However, traditional power modeling techniques fall short of meeting the requirements. The existing standard VLSI design flow generates accurate power evaluations through multiple design stages, including RTL implementation, logic synthesis, RTL simulation with realistic workloads, and gate-level power simulation using commercial tools~\cite{powerpro, ptpx}. Unfortunately, this process is excessively time-consuming for evaluating each architectural-level design configuration. As for faster alternatives, widely-adopted architectural-level power models such as McPAT~\cite{li2009mcpat,tang2014mcpat,guler2020mcpat}, and Wattch~\cite{brooks2000wattch} have been criticized for their unreliable accuracy, as discussed in many prior studies~\cite{xi2015quantifying, nowatzki2015architectural}. Despite some improvements in subsequent works, they are primarily developed in-house to cater to proprietary designs~\cite{xi2015quantifying}.

In recent years, some ML methods~\cite{zhai2022mcpat, zhai2023microarchitecture, lee2015powertrain} were proposed to directly calibrate the existing analytical models like McPAT~\cite{li2009mcpat} (i.e., using McPAT output as ML models' input). In this way, they may generate a more accurate estimation, when the target design architecture is similar to already known designs in the training dataset. However, their accuracies degrade significantly when applied to unknown new design configurations. This problem is particularly severe when training data is limited, which is frequently the case in practice. As mentioned, the label collection of a new sample (i.e., new architecture-level configuration) requires the actual implementation of it through VLSI flow and workload-based simulations. This whole process can be extremely time-consuming. A most recent ML work~\cite{zhai2023microarchitecture} proposes to predict unknown new designs with transfer learning. But it still requires a few ground-truth samples in the target configuration domain, which is still time-consuming to generate in practice. In addition, some design space exploration (DSE) works~\cite{lee2007illustrative, bai2021boom} also develop their own ML-based power models, which are trained iteratively based on labels collected during exploration. Besides still being limited by training data, these models also typically cannot incorporate workload-related information, thus \textcolor{black}{failing} to predict vector-based power values for each specific workload.

\begin{figure}[!t]
\centering
\includegraphics[width=0.41\textwidth]{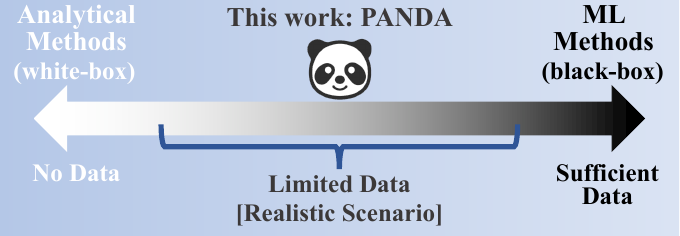}
\caption{A summary of architecture-level power modeling methods. Traditional analytical methods adopt inaccurate oversimplified handcrafted model, while the accuracies of ML methods degrade significantly when applied to unknown new configurations. PANDA unifies both analytical and ML solutions, addressing their long-lasting limitations.}
\vspace{-.2in}
\label{summary_panda}
\end{figure}

\begin{table*}[!t]
      \centering
      \vspace{-.1in}
      \renewcommand{\arraystretch}{1.1}
      \resizebox{\textwidth}{!}{
        \begin{tabular}{ |c|c|c|c| } 
        \hline
        Component $i$ &  Configuration parameters $C_i$ of each component  &  Event parameters $E_i$ of each component & CPU part  \\
        \hline
         \hline
BP &  FetchWidth, BranchCount  & BTBLookups, condPredicted, condIncorrect, commit.branches & \multirow{8}{*}{Frontend} \\
         \cline{1-3}
\multirow{4}{*}{IFU} &  \multirow{3}{*}{FetchWidth, DecodeWidth}   & fetch.insts, fetch.branches, fetch.cycles, numRefs, numStoreInsts, numInsts,  &   \\
         &           & decode.runCycles, decode.blockedCycles, decode.decodedInsts, numBranches, & \\
         &  FetchBufferEntry, ICacheFetchBytes  & intInstQueueReads, intInstQueueWrites, intInstQueueWakeupAccesses, & \\ 
         & & fpInstQueueReads, fpInstQueueWrites, fpInstQueueWakeupAccesses & \\
         \cline{1-3}
I-TLB &  ICacheTLBEntry  & itb.accesses, itb.misses &  \\
         \cline{1-3}
\multirow{2}{*}{I-Cache} &  \multirow{2}{*}{ICacheWay, ICacheFetchBytes} & icache.overallAccesses, icache.overallMisses, icache.ReadReq.mshrHits, & \\  
& & icache.ReadReq.mshrMisses, icache.tagAccesses & \\
         \hline
         \hline

RNU & DecodeWidth  & intLookups, renamedOperands, fpLookups, renamedInsts, runCycles, blockCycles, committedMaps  & \multirow{6}{*}{Execution}\\
         \cline{1-3}
ROB &  DecodeWidth, RobEntry  & rob.reads, rob.writes & \\
         \cline{1-3}
\multirow{2}{*}{ISU} &  DecodeWidth, MemIssueWidth,   & IssuedMemRead, IssuedMemWrite, IssuedFloatMemRead, IssuedFloatMemWrite, &  \\
             &  FpIssueWidth, IntIssueWidth  & IssuedIntAlu, IssuedIntMult, IssuedIntDiv, IssuedFloatMult, IssuedFloatDiv & \\
        \cline{1-3}
Regfile &  DecodeWidth, IntPhyRegister, FpPhyRegister  & intRegfileReads, fpRegfileReads, intRegfileWrites, fpRegfileWrites, functionCalls &   \\
         \cline{1-3}
FU Pool &  MemIssueWidth, FpIssueWidth, IntIssueWidth  & intAluAccesses, fpAluAccesses &  \\
        \hline
        \hline

LSU &  LDQEntry, STQEntry, MemIssueWidth  & MemRead, InstPrefetch, MemWrite & \multirow{4}{*}{Mem Access} \\
         \cline{1-3} 
D-TLB &  DCacheTLBEntry  & dtb.accesses, dtb.misses & \\
         \cline{1-3}
\multirow{2}{*}{D-Cache} &  DCacheWay, DCacheTLBEntry,  & dcache.ReadReq.accesses, dcache.WriteReq.accesses, dcache.ReadReq.misses, dcache.WriteReq.misses, &  \\
         & DCacheMSHR, MemIssueWidth  & dcache.overallMisses, dcache.MshrHits, dcache.MshrMisses, dcache.tagAccesses & \\
        \hline
        \hline
         Other Logic &  All  & All & Other Logic\\
        \hline
        \end{tabular}
        }
        \caption{Our identified architecture-level design configuration parameters $C_i$ and event parameters $E_i$ of each $i^\text{th}$ component.}
        \label{tbl:config_event}
        \vspace{-.1in}
\end{table*}

In this work, we first propose our qualitative analysis of both analytical~\cite{li2009mcpat, brooks2000wattch} and ML-based~\cite{zhai2022mcpat, zhai2023microarchitecture, lee2015powertrain, lee2007illustrative, bai2021boom} architecture-level models, as summarized in Fig.~\ref{summary_panda}. 
When there is sufficient training data that can cover the potential testing data the model will encounter, the ML model performs better. However, when very limited data is available, ML models can even be misleading, and traditional analytical methods become the only option. Nevertheless, in realistic scenarios, most design teams are somewhere in the middle, with a very limited number of already-implemented architecture configurations that can be used as training data. The high demand for the training data amount and diversity sets a high barrier \textcolor{black}{to} the wide adoption of ML solutions~\cite{zhai2022mcpat, zhai2023microarchitecture, lee2015powertrain, lee2007illustrative, bai2021boom} in practice. 


Inspired by these observations, we propose an innovative new architectural-level power modeling solution named PANDA. As Fig.~\ref{summary_panda} shows, its name PANDA implies unifying \emph{white}-box analytical models and \emph{black}-box ML models\footnote{Please notice that most existing ML works~\cite{zhai2022mcpat, zhai2023microarchitecture, lee2015powertrain, lee2007illustrative, bai2021boom} directly adopt analytical model McPAT's output as an input feature and perform calibration with ML. But for the main part of their method, unlike analytical solutions, they do not explicitly capture/encode any patterns based on architecture knowledge. This is essentially different from how PANDA integrates analytical model, and thus we categorize these prior works~\cite{zhai2022mcpat, zhai2023microarchitecture, lee2015powertrain, lee2007illustrative, bai2021boom} as ML methods in Fig.~\ref{summary_panda}.}, benefiting from the complementary advantages of both sides. PANDA adopts an analytical framework to model the hierarchy of individual components. For each component, it integrates an ML model with a simple customized analytical function, which is based on identified key configuration parameters of this component. In this way, the analytical function captures the key pattern of the component, leaving more complex detailed patterns to be learned by the ML portion. As a result, PANDA significantly outperforms state-of-the-art solutions, especially when training data is very limited. The low requirement on training data \textcolor{black}{makes} PANDA readily to be widely adopted. In addition, unlike most existing ML methods~\cite{zhai2022mcpat, zhai2023microarchitecture, lee2015powertrain, lee2007illustrative, bai2021boom} that rely on existing analytical models like McPAT~\cite{li2009mcpat} as model input \textcolor{black}{features} and simply perform calibration, PANDA is a standalone solution without relying on any existing analytical models.



Besides power modeling, to comprehensively evaluate a design at the \textcolor{black}{architecture level}, evaluations of other design qualities like performance (i.e., number of cycles to complete a workload), area, and energy are also desired. For performance, cycle-level simulators like gem5~\cite{binkert2011gem5} are widely-adopted, but it is not sufficiently accurate. For gate area, McPAT~\cite{li2009mcpat} is widely-adopted but inaccurate. 
To solve these challenges, our solution PANDA also enables accurate evaluations of other design objectives.

The key contributions in this work can be summarized below. 
\begin{itemize} 
    \item We analyzed the root cause of limited accuracy in both analytical and ML-based power models, then propose an open-sourced architecture-level power modeling solution named PANDA\footnote{It has been open-sourced at https://github.com/zqj2333/PANDA}, which unifies these two major types of methods. Given different training dataset \textcolor{black}{sizes}, it outperforms state-of-the-art baselines by 5\% to 30\% in error percentage. The gap is increasingly obvious as the training data amount decreases. 
    \item In this work, we provide a unified formulation that can express all existing architecture-level power models including PANDA. It helps to demonstrate the portion of analytical and ML techniques in each method and guide the accuracy analysis when training data amount varies. 
    \item Unlike most ML methods that directly calibrate McPAT, PANDA does not rely on any existing analytical power model like McPAT to provide features. Instead, PANDA develops its own simpler analytical function for each component based on architecture knowledge, leaving more complex patterns to be learned by the ML part. Such a standalone solution avoids propagating the unreliable accuracies of McPAT. 
    \item Besides power modeling, PANDA also models other design objectives including design performance, area, and energy. 
    \item Finally, PANDA further supports the power prediction targeting unknown new technology nodes. 
\end{itemize}





%% file: _txt/4_Methodology.tex
\section{Methodology} \label{sec:algorithm}


\begin{figure}[!b]
\centering
\vspace{-.15in}
\includegraphics[width=0.4\textwidth]{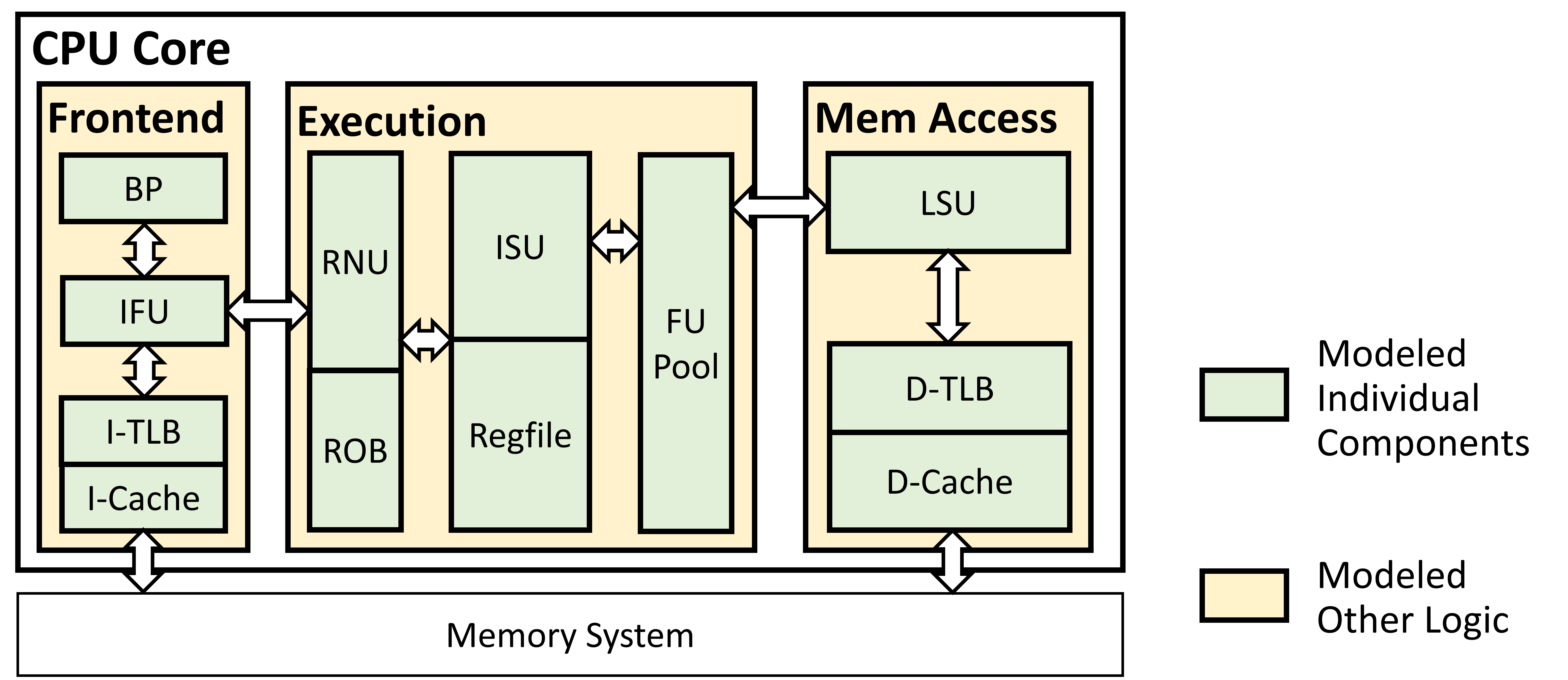}
\caption{The architecture of our target Out-of-Order RISC-V CPU core. The green blocks indicate key individual components modeled by PANDA. These components correspond to the Table~\ref{tbl:config_event}. The yellow block refers to the Other Logic indicated in Table~\ref{tbl:config_event}.}
\label{arch}
\end{figure}

\begin{figure*}[!t]
\centering
\vspace{-.1in}
\subfigure[Analytical power model]{
    \centering
    \includegraphics[height=0.20\textwidth]{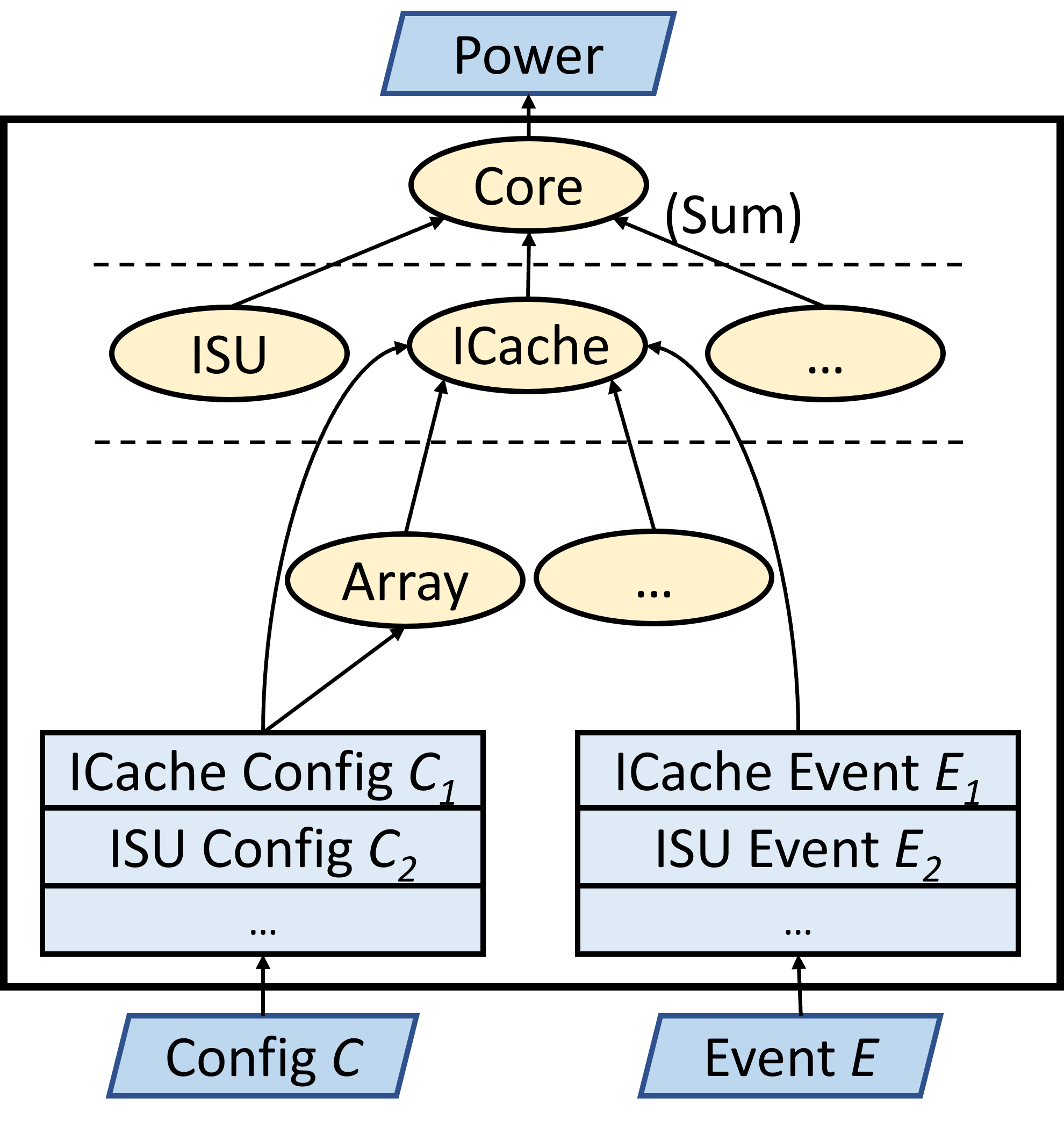}
    \label{abstract0}
}\hspace{.2in}
\subfigure[ML-based power model]{
    \centering
    \includegraphics[height=0.20\textwidth]{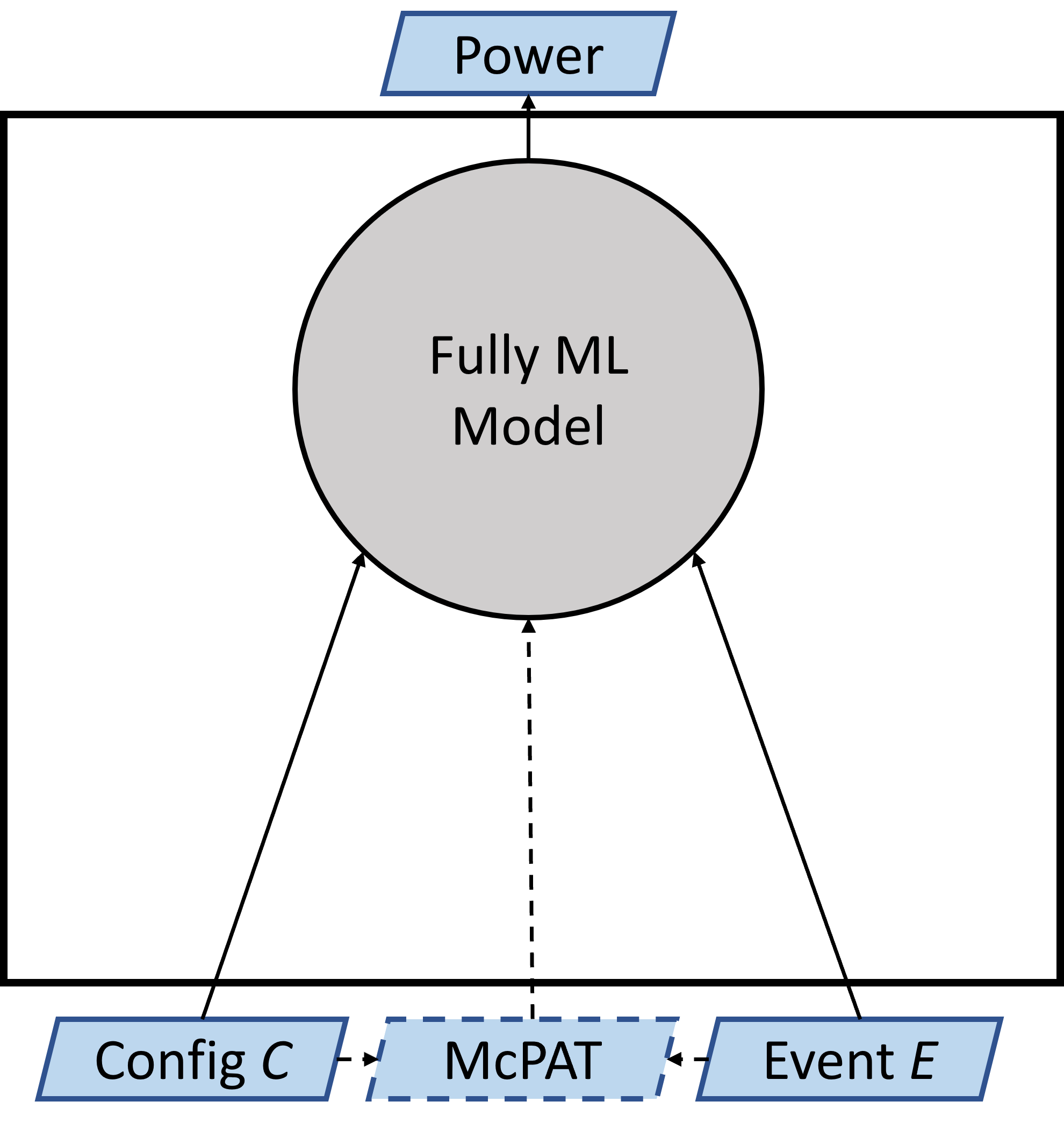}
    \label{abstract1p5}
}\hspace{.2in}
\subfigure[Component-level ML model]{
    \centering
    \includegraphics[height=0.20\textwidth]{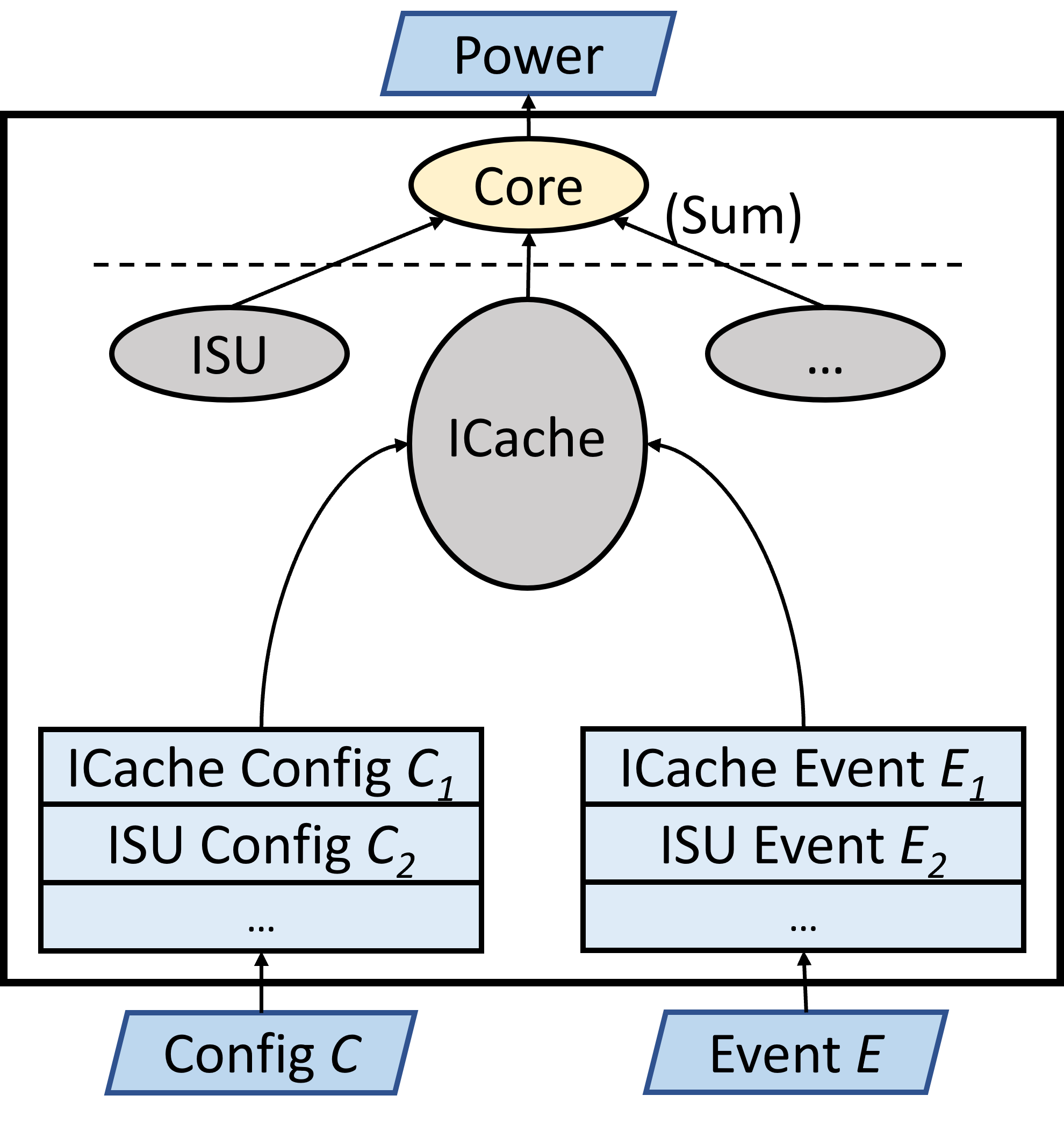}
    \label{abstract2}
}\hspace{.2in}
\subfigure[PANDA]{
    \centering
    \includegraphics[height=0.20\textwidth]{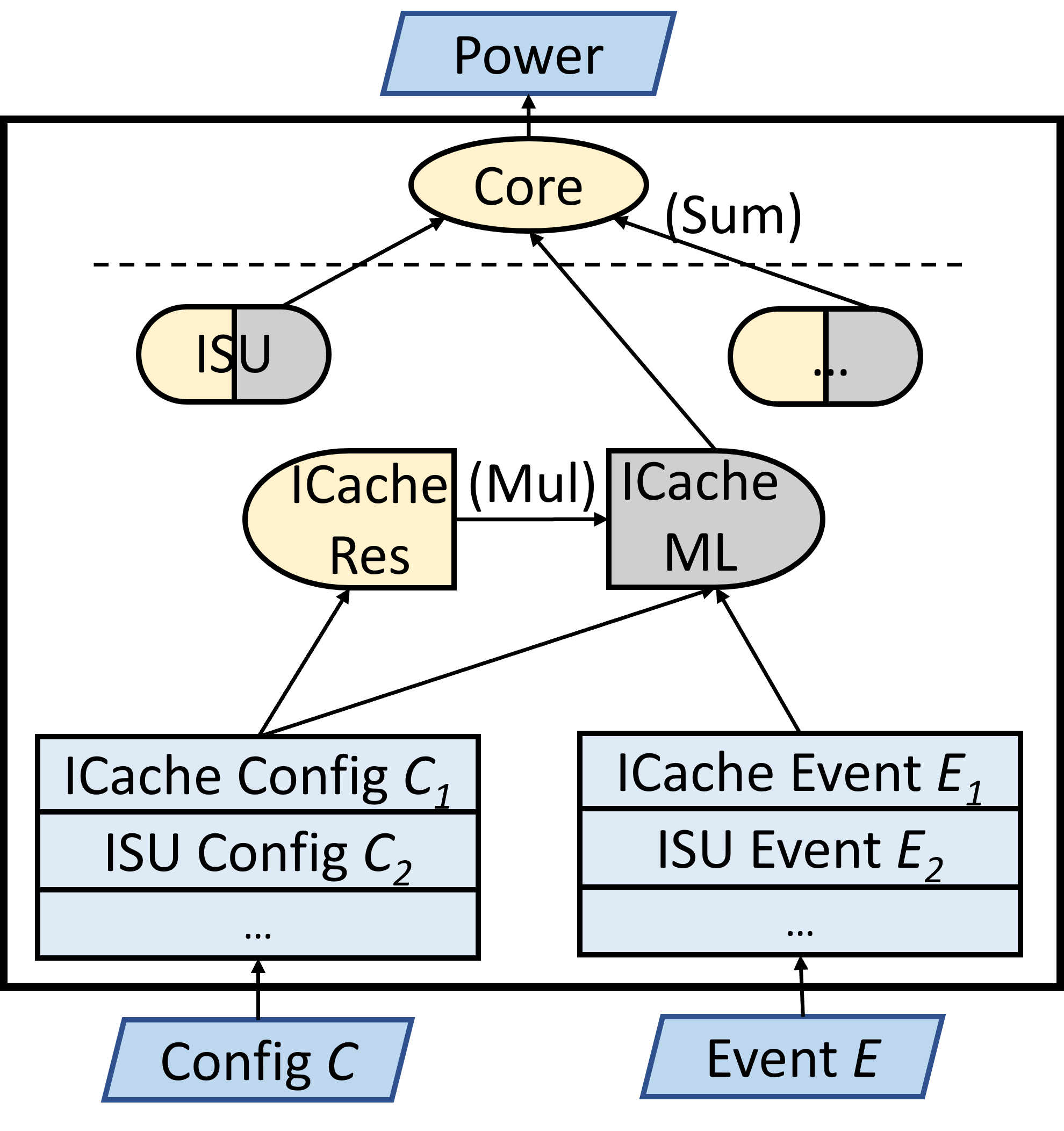}
    \label{abstract3}
}
\vspace{-.1in}
\caption{Illustration of different power modeling methods, yellow means the analytical parts and dark means the ML parts. (a) The analytical method~\cite{li2009mcpat,brooks2000wattch}. (b) The ML method~\cite{zhai2022mcpat, zhai2023microarchitecture, lee2015powertrain, lee2007illustrative, bai2021boom}, the method mainly relies on ML model. (c) Component-level power model, a much weaker variant of PANDA for ablation study. It builds an ML model to predict the power of each component. (d) PANDA, the model of each component consists of two sub-model, the yellow one means the resource function, which is analytical, while the dark one means the ML model.}
\label{abstract_model}
\vspace{-.16in}
\end{figure*}

\subsection{CPU Components and Configuration Parameters Identification}


For our target modern out-of-order (OoO) CPU core\footnote{PANDA \textcolor{black}{experiments} on the RISC-V OoO CPU core BOOM~\cite{zhao2020sonicboom}. It can be extended to other CPU types with minor modifications.}, all major architecture-level configuration \textcolor{black}{parameters} and event parameters explored in this work \textcolor{black}{are} listed in Table~\ref{tbl:config_event}. Denote all these configuration parameters as a set $C$ and all workload-related architecture-level event parameters as $E$. To develop PANDA, we further identified key CPU components that can be individually modeled in power evaluation. Table~\ref{tbl:config_event} lists our identified configuration \textcolor{black}{parameters} and event parameters related \textcolor{black}{to} each component, and Fig.~\ref{arch} shows the overall CPU architecture including these components.


As the CPU core architecture in Fig.~\ref{arch} shows, all our modeled individual components can be categorized into three main parts: frontend, execution, and memory access, with each part comprised of several key components, as introduced below.
\begin{itemize}
    \item The CPU frontend part includes branch predictor (BP), instruction fetch unit (IFU), instruction translation-lookaside buffer (I-TLB), and L1 instruction cache (I-Cache). 
    \item The CPU execution part includes rename unit (RNU), reorder buffer (ROB),  issue unit (ISU), register file (Regfile), and functional unit pool (FU Pool), which includes the ALUs, floating-point units, and other functional units.
    \item The CPU memory-access part includes data translation-lookaside buffer (D-TLB), data cache (D-Cache), and the remaining logic in load/store unit (LSU). 
\end{itemize}
All other CPU design logic not covered by above components is referred to as Other Logic.

Generally, \textcolor{black}{assuming} there are $N$ components in the target CPU design, our identified configuration \textcolor{black}{parameters} related \textcolor{black}{to} the $i^{\text{th}}$ component \textcolor{black}{are} denoted as $C_i$, with $C = \{C_i \, | \, i\in [1, N]\}$. Similarly, denote the architecture-level event parameters related \textcolor{black}{to} this component as $E_i$, with $E = \{E_i \, | \, i\in [1, N]\}$. Both $C_i$ and $E_i$ of each component can be simply looked up in Table~\ref{tbl:config_event}.



\subsection{A General Formulation of Existing Power Modeling Methods}



Fig.~\ref{abstract_model} further provides conceptual visualizations of existing analytical power models~\cite{li2009mcpat, brooks2000wattch}, ML-based models~\cite{zhai2022mcpat, zhai2023microarchitecture, lee2015powertrain}, and PANDA, under a similar framework, with all CPU configuration \textcolor{black}{parameters} $C$ and workload-related event parameters $E$ as model input candidates. We find that power modeling works can be expressed as a unified formulation. So we first propose our formulation of the two types of prior works, then we will start introducing our new method in the next Subsection.

\textbf{Formulation of ML works.} Existing ML solutions~\cite{zhai2022mcpat, zhai2023microarchitecture, lee2015powertrain} simply build ML models targeting total power values, based on all available design configuration parameters $C$ and event parameters $E$, as shown below\footnote{For simplicity, we do not include the McPAT output as a potential input feature in the formulation of ML works.}. The $\boldsymbol{F_{ml}}$ denotes data-driven ML methods, including data-driven feature selection and the development of ML models. It can be formulated below, with $P_{\text{ml}}$ denoting the power prediction value.
\begin{equation*}
P_{\text{ml}} = \boldsymbol{F_{ml}} \; (\{C, E\}) 
\end{equation*}
It can be rewritten as an equivalent general form below by explicitly indicating configuration parameters of all components.  
\begin{equation}
P_{\text{ml}} = \boldsymbol{F_{ml}} \; (\{C_i \, | \, i\in [1, N]\}, \; \{E_i  \, | \, i\in [1, N]\}) \label{eq2}
\end{equation}
It means existing ML methods adopt the available configuration parameters and event parameters information from all components to evaluate the total power value of the whole design.

\textbf{Formulation of analytical works.} Different from ML methods, analytical methods like McPAT~\cite{li2009mcpat} explicitly design separate analytical models for each component, whose estimated power is denoted as $P_{\text{ana}}^i$, according to designers' background knowledge. We formulate such analytical methods for each component $i$ as below,
\begin{align}
P_{\text{ana}}^i = \boldsymbol{F_{agg}}^i \;& (E_i, \; \boldsymbol{F_{res}}^i (C_i))  \label{eq:ana_i} 
\end{align}
where an analytical `resource function' $\boldsymbol{F_{res}}^i(C_i)$ first calculates an intermediate value that reflects the resource consumption corresponding to the configuration parameters. 
Then another `aggregation function' $\boldsymbol{F_{agg}}^i$ \textcolor{black}{combines} both resource values $\boldsymbol{F_{res}}^i(C_i)$ and event parameters $E_i$ to calculate the power of this component for each workload.




We illustrate the aforementioned analytical methods with the actual I-Cache component in CPU frontend as an example. The configuration parameters $C_i$ of I-Cache include the number of ways of the N-set associated cache (i.e., \emph{ICacheWay}) and the unit of line capacity that I-Cache supports (i.e., \emph{ICacheFetchBytes}). Analytical models like McPAT compute the power consumption based on the number of \textcolor{black}{hits and misses}. We can formulate its resource function $\boldsymbol{F_{res}}^i$ as estimating the energy per hit and miss based on the I-Cache configuration parameters.
\begin{equation}
     \boldsymbol{F_{res}}^i\; (\textit{ICacheWay}, \; \textit{ICacheFetchBytes}) = \textit{Energy per hit/miss}
\end{equation}
Then aggregation function $\boldsymbol{F_{agg}}^i$ combines the resources and corresponding event parameters, including \textcolor{black}{the} number of hits and \textcolor{black}{misses}. Then the actual implementation of Equation~(\ref{eq:ana_i}) for I-Cache component can be expressed as below.  
\begin{align*}
    P_{\text{ana}}^i &= \boldsymbol{F_{agg}}^i \; (\textit{\#Hit, \#Miss, Energy per hit/miss}) \\[2pt]
    &= \frac{\textit{\#Hit} * \textit{Energy per hit} + \textit{\#Miss} * \textit{Energy per miss}}{\textit{Total benchmark execution time}} 
\end{align*}

Based on predicted component power, the total power is simply the summation \textcolor{black}{of} all components $P_{\text{ana}} = \sum_{i \in [1, N]} P_{\text{ana}}^i$.

\subsection{The General Formulation of PANDA}

\textbf{Formulation of PANDA.} In contrast with ML methods in Equation~(\ref{eq2}) and analytical methods in Equation~(\ref{eq:ana_i}), we combine the advantages of both methods in this work. The general expression for each component $i$ is shown below. 
\begin{equation*}
P^i_{\text{PANDA}} = \boldsymbol{F_{agg}}^i\; ( \boldsymbol{F_{ml}}^i (C_i, E_i), \; \boldsymbol{F_{res}}^i (C_i)) \label{eq:gem}
\end{equation*}


Similar to the notations used in Equation~(\ref{eq2})(\ref{eq:ana_i}), here the $\boldsymbol{F_{agg}}^i$ and $\boldsymbol{F_{res}}^i$ denote analytical functions, and $\boldsymbol{F_{ml}}^i$ denotes an ML-based function\footnote{Please notice that the actual functions of $\boldsymbol{F_{agg}}^i$, $\boldsymbol{F_{res}}^i$, $\boldsymbol{F_{ml}}^i$ in PANDA are different from Equation~(\ref{eq2})(\ref{eq:ana_i}). We use \textcolor{black}{the} same notation to simplify the expression and demonstrate the general form of our formulation.}. The general formulation combines the ML model in Equation~(\ref{eq2}) and the analytical model in Equation~(\ref{eq:ana_i}). Now we start to introduce each part and explain its advantages.

First, we design an analytical resource function $\boldsymbol{F_{res}}^i(C_i)$ 
according to background knowledge of how the configuration \textcolor{black}{parameters} $C_i$ will affect the power of this component. Compared with the similar function in Equation~(\ref{eq:ana_i}), we capture the simpler yet primary pattern in this function, and leave complex patterns to be learned by our ML model. 




Using the same I-Cache component example, for a typical N set-associative I-Cache, each cache access requires accessing both tag and data array in all cache ways simultaneously for lower latency. It causes the power consumption \textcolor{black}{to be} roughly proportional to the number of cache ways (i.e., \emph{ICacheWay}). Regarding the \emph{ICacheFetchBytes}, it decides the size of the cache line of the I-Cache, so the power of accessing a cache line in a way will scale proportionally with it. Considering both factors, our resource function is as below. 
\begin{align}
    \boldsymbol{F_{res}}^i (C_i) &= \textit{ICacheWay} *  \textit{ICacheFetchBytes}  \label{eq:icache} 
\end{align}

Second, we propose the ML model $\boldsymbol{F_{ml}}^i(C_i, E_i)$ based on both configuration parameters and event parameters for each component $i$. It learns all the detailed correlations beyond the simple correlation in resource function. Finally, the estimations of ML model $\boldsymbol{F_{ml}}^i$ and resource function $\boldsymbol{F_{res}}^i$ are multiplied to generate the final power estimation. The finalized PANDA formulation is shown below. 
\begin{align}
    P_{\text{PANDA}}^i &= \boldsymbol{F_{agg}}^i\; ( \boldsymbol{F_{ml}}^i (C_i, E_i), \; \boldsymbol{F_{res}}^i (C_i)) \nonumber \\[2pt]
            &= \boldsymbol{F_{ml}}^i (C_i, E_i) * \boldsymbol{F_{res}}^i\; (C_i) \label{eq:panda}
\end{align}

As Equation~(\ref{eq:panda}) shows, PANDA adopts a simple multiplication to aggregate the ML model $\boldsymbol{F_{ml}}^i$ and analytical resource function $\boldsymbol{F_{res}}^i$. Although there may not be a definite answer, here we share three intuitive thoughts behind this decision. 1) As the I-Cache example in Equation~\ref{eq:icache} shows, our identified resource function for each component is theoretically roughly proportional to the power consumption. Therefore, there should be a linear relationship between $\boldsymbol{F_{res}}^i$ and power prediction, multiplication is obviously the simplest option. 2) A possible alternative is to directly incorporate the resource function $\boldsymbol{F_{res}}^i$ as an input feature of ML model $\boldsymbol{F_{ml}}^i$. But this is not a good option. Just like existing ML solutions calibrating McPAT, the ML model accuracy degrades significantly when training data is limited. When only used as ML model input, the strength of the analytical resource function is not fully utilized to tackle the data availability problem. 3) Another interesting perspective will be provided by the analysis in Section~\ref{sec:resource_ana}, it demonstrates that such multiplication in Equation~(\ref{eq:panda}) makes ML model learn to predict power/$F_{res}$, whose distribution across different configurations is much more uniform than the power value alone. The result implies that ML model should learn power/$F_{res}$ better when the training data is limited. Combining these three reasons, we believe multiplication is the correct option, and it indeed provides excellent performance. 


Substituting the resource function $\boldsymbol{F_{res}}^i\; (C_i)$ in Equation~(\ref{eq:panda}) with Equation (\ref{eq:icache}), the power of the I-Cache example is shown below.
\begin{equation}
    P_{\text{PANDA}}^i = \boldsymbol{F_{ml}}^i (C_i, E_i) * \textit{ICacheWay} *  \textit{ICacheFetchBytes}
\end{equation}

\begin{table}[!t]
      \centering
      \vspace{.05in}
     \hspace{-.1in}
      \renewcommand{\arraystretch}{1.1}
      \resizebox{0.5\textwidth}{!}{
        \begin{tabular}{ |c|c||c|c| } 
        \hline
        Component   &  $\boldsymbol{F_{res}}^i$ & Component & $\boldsymbol{F_{res}}^i$   \\
        \hline
         \hline
         I-Cache &  ICacheWay * ICacheFetchBytes & BP &  FetchWidth \\
         \hline
         ISU &  $f_{\text{ReserveStationNum}}$(DecodeWidth)  & IFU &  DecodeWidth \\
         \hline
         Regfile &  IntPhyRegister + FpPhyRegister & I-TLB &  ICacheTLBEntry + bias \\
         \hline
         D-Cache &  DCacheWay * MemIssueWidth & RNU &  DecodeWidth \\
         \hline
         LSU &  LDQEntry + STQEntry & ROB &  RobEntry  \\
         \hline
         D-TLB &  DCacheTLBEntry + bias & FU Pool &  1 \\
         \hline
         Other Logic &  DecodeWidth + bias \\
         \cline{1-2}
        \end{tabular}
        }
        \caption{PANDA's resource function $\boldsymbol{F_{res}}^i$ of each major component $i$ in the target out-of-order CPU core. They are derived based on the background knowledge of CPU architecture design.}
        \label{resource_table}
        \vspace{-.15in}
\end{table}


\subsection{Resource Functions and ML Model of PANDA}

Table~\ref{resource_table} shows our proposed key resource functions\footnote{In the remaining of this paper, we may directly denote the resource function $\boldsymbol{F_{res}}^i(C_i)$ as $\boldsymbol{F_{res}}^i$ for simplicity.} $\boldsymbol{F_{res}}^i$ for all major components in the target out-of-order CPU core. We introduce how we derive these functions below.

Similar to the aforementioned I-Cache example, for an N-way set-associative L1 data cache (D-Cache), the power is roughly proportional to the number of cache ways (i.e., \emph{DCacheWay}). In addition, modern CPUs support serving multiple read requests simultaneously to improve throughput~\cite{zhao2020sonicboom}. The power consumption is proportional to the number of simultaneously accessed cache lines. This simultaneous cache access number typically equals the number of memory-access instructions issued each time (i.e., \emph{MemIssueWidth}). Therefore, we propose $\boldsymbol{F_{res}}^i$ = \emph{DCacheWay} * \emph{MemIssueWidth}.

\circled{1} \textbf{Frontend.} The Frontend part of the Out-of-Order CPU consists of 4 main components, including BP, IFU, I-TLB, and I-Cache. The I-Cache has been discussed.
(1) The branch predictor (BP) is one of the most crucial parts of the frontend of modern OoO CPUs. It has a significant impact on CPU performance. 
A more accurate BP requires larger SRAM-based or register-based tables like the branch history table, leading to higher power consumption. In many modern CPUs, the size of the branch predictor is designed to scale proportionally with the number of instructions being fetched each time (i.e., \emph{FetchWidth}) at the frontend. Therefore, we set $\boldsymbol{F_{res}}^i$ = \emph{FetchWidth} as the BP's resource function.
(2) For Instruction Fetch Unit (IFU), the size of its instruction fetch buffer and decode logic depends on the \emph{DecodeWidth}. In comparison, the impact \textcolor{black}{of} other components on the IFU is secondary. Therefore, we set $\boldsymbol{F_{res}}^i$ = \emph{DecodeWidth} for IFU. 
(3) For the I-TLB, its power is mainly affected by the number of TLBEntry (i.e., \emph{ICacheTLBEntry}), but there is also a part that \textcolor{black}{remains} unchanged when increasing the number of TLBEntry. Therefore, we set $\boldsymbol{F_{res}}^i$ = \emph{ICacheTLBEntry} + bias for I-TLB, with the bias denoting a constant power value. Note that the bias will be readily estimated by fitting this linear function on training data, as long as there are more than one training samples. 

\circled{2} \textbf{Execution.} The Execution part of the Out-of-Order CPU consists of 5 main components, including RNU, ROB, ISU, Regfile, and FU Pool.
(1) For the renaming unit (RNU), the RNU typically has a renaming width equal to the \emph{DecodeWidth}. Therefore, we set $\boldsymbol{F_{res}}^i$ = \emph{DecodeWidth} for RNU.
(2) The reorder buffer (ROB), the power consumption mainly depends on the size of it, so the power consumption is proportional to the number of rob \textcolor{black}{entries} (i.e., \emph{RobEntry}). Therefore, we set $\boldsymbol{F_{res}}^i$ = \emph{RobEntry} for ROB.
(3) The issue unit (ISU) is a critical part of OoO CPUs as it manages instruction scheduling and pipeline information. The reserve stations, which store the scheduling information, are the main component of ISU and largely affect the ISU power consumption. The number of reserve station entries typically depends on the number of instructions being decoded each time (i.e., DecodeWidth). 
We set $\boldsymbol{F_{res}}^i$ = $f_{\text{ReserveStationNum}}$(\emph{DecodeWidth}), where the $f_{\text{ReserveStationNum}}$ is a look-up table that maps \emph{DecodeWidth} to the number of reserve stations. It will be available as part of the CPU design.
(4) The Regfile is mainly comprised of integer physical registers and float physical registers. So the power is proportional to the sum of the integer physical register size and the float physical register size. Therefore, we set $\boldsymbol{F_{res}}^i$ = \emph{IntPhyRegister} + \emph{FpPhyRegister} for Regfile.
(5) The FU Pool is very complex, comprised of a variety of function units, and each function unit has different power \textcolor{black}{characteristics}. Therefore, we set $\boldsymbol{F_{res}}^i$ = 1 for FU Pool, handing this over to the ML function.

\circled{3} \textbf{Memory Access.} The Memory Access part of the Out-of-Order CPU consists of 3 main components, including LSU, D-TLB, and D-Cache, the D-Cache has been discussed.
(1) The load store unit (LSU) is mainly comprised of the load queue and store queue. Its power consumption is closely tied to the total number of entries in these queues, denoted as \emph{LDQEntry} + \emph{STQEntry}. Therefore, we set $\boldsymbol{F_{res}}^i$ = \emph{LDQEntry} + \emph{STQEntry} for LSU.
(2) For the D-TLB, this component is very similar to I-TLB. Similarly, we set $\boldsymbol{F_{res}}^i$ = \emph{DCacheTLBEntry} + bias for D-TLB.

\circled{4} \textbf{Other Logic.} `Other Logic' is the most complex part of the CPU, consisting \textcolor{black}{of} all other control and pipeline logic except existing components. It may seem that the power consumption of `Other Logic' is difficult to estimate, but we have identified \emph{DecodeWidth} as a useful indicator. \emph{DecodeWidth} determines the number of instructions that can be decoded simultaneously, and it typically also equals the width of several subsequent pipeline stages, such as the integer rename width, ROB width, and commit width, etc. Therefore, \emph{DecodeWidth} can be referred to as the general \emph{pipeline width} of the whole CPU design. A large portion of the control logic and pipeline logic scale with this pipeline width, while other parts remain unchanged even as other components of the CPU scale out. Therefore, we derive the resource function as $\boldsymbol{F_{res}}^i$ = \emph{DecodeWidth} + bias, with the bias denoting a constant power value. Similar to I-TLB, the bias can be estimated using training data. 

As for the ML model $\boldsymbol{F_{ml}}^i$ of each component, we all adopt Gradient Boosting Trees ~\cite{friedman2001greedy} like XGBoost ~\cite{chen2016xgboost}, a widely-adopted machine learning algorithm for tabular data type, to build regressors. To avoid introducing our engineers' bias during ML model hyper-parameter tuning, for all these XGBoost ML models in PANDA, we simply adopt the default hyper-parameters (i.e., max depth=6, num of estimatros=100) without any further parameter tuning. PANDA is already sufficiently accurate in this case.

\begin{table*}[!t]
\centering
      \vspace{-.05in}
      \resizebox{0.92\textwidth}{!}{
        \begin{tabular}{ |c||c c c|c c c|c c c|c c c|c c c||c c| } 
\hline
Configuration Parameter  & C1 & C2 & C3 & C4 & C5 & C6 & C7 & C8 & C9 & C10 & C11 & C12 & C13 & C14 & C15 & SP1 & SP2\\
\hline
\hline
FetchWidth & 4 & 4 & 4 & 4 & 4 & 8 & 8 & 8 & 8 & 8 & 8 & 8 & 8 & 8 & 8 & 8 & 8\\
\hline
DecodeWidth & 1 & 1 & 1 & 2 & 2 & 2 & 3 & 3 & 3 & 4 & 4 & 4 & 5 & 5 & 5 & 1 & 5\\
\hline
FetchBufferEntry & 5 & 8 & 16 & 8 & 16 & 24 & 18 & 24 & 30 & 24 & 32 & 40 & 30 & 35 & 40 & 10 & 40\\
\hline
RobEntry & 16 & 32 & 48 & 64 & 64 & 80 & 81 & 96 & 114 & 112 & 128 & 136 & 125 & 130 & 140 & 16 & 140\\
\hline
IntPhyRegister & 36 & 53 & 68 & 64 & 80 & 88 & 88 & 110 & 112 & 108 & 128 & 136 & 108 & 128 & 140 & 36 & 140\\
\hline
FpPhyRegister & 36 & 48 & 56 & 56 & 64 & 72 & 88 & 96 & 112 & 108 & 128 & 136 & 108 & 128 & 140 & 36 & 140\\
\hline
LDQ/STQEntry & 4 & 8 & 16 & 12 & 16 & 20 & 16 & 24 & 32 & 24 & 32 & 36 & 24 & 32 & 36 & 4 & 36\\
\hline
BranchCount & 6 & 8 & 10 & 10 & 12 & 14 & 14 & 16 & 16 & 18 & 20 & 20 & 18 & 20 & 20 & 6 & 20\\
\hline
MemIssue/FpIssueWidth & 1 & 1 & 1 & 1 & 1 & 1 & 1 & 1 & 2 & 1 & 2 & 2 & 2 & 2 & 2 & 1 & 2\\
\hline
IntIssueWidth & 1 & 1 & 1 & 1 & 2 & 2 & 2 & 3 & 3 & 4 & 4 & 4 & 5 & 5 & 5 & 1 & 5\\
\hline
DCache/ICacheWay & 2 & 4 & 8 & 4 & 4 & 8 & 8 & 8 & 8 & 8 & 8 & 8 & 8 & 8 & 8 & 2 & 2\\
\hline
DTLBEntry & 8 & 8 & 16 & 8 & 8 & 16 & 16 & 16 & 32 & 32 & 32 & 32 & 32 & 32 & 32 & 8 & 32\\
\hline
DCacheMSHR & 2 & 2 & 4 & 2 & 2 & 4 & 4 & 4 & 4 & 4 & 4 & 8 & 8 & 8 & 8 & 2 & 8\\
\hline
ICacheFetchBytes & 2 & 2 & 2 & 2 & 2 & 4 & 4 & 4 & 4 & 4 & 4 & 4 & 4 & 4 & 4 & 4 & 4\\
         \hline
        \end{tabular}
        }
        \caption{The configurations that we used in our experiment. The first 15 configurations (named C1-C15) are normal configurations, which is divided into 5 domains depending on the DecodeWidth, the last two configuration (named SP1, SP2) are special case that will be used in the case study part. To be specific, the SP1 is derived from C1, which is a CPU with a large BP but small other components, the SP2 is derived from C15, which is a CPU with a small D-Cache and I-Cache but large other components.}
        \label{configtable}
        \vspace{-.1in}
\end{table*}

\subsection{Other Design Quality Prediction}



Besides power prediction, PANDA also supports evaluation of other design qualities including area, performance (i.e., \textcolor{black}{the} number of \textcolor{black}{cycles} to complete a given workload), and energy. (1) The area model is similar to aforementioned component-level power model, but only uses configuration parameters $C_i$ as features, without including workload event parameters $E_i$. (2) For performance prediction, we observe that gem5~\cite{binkert2011gem5} achieves a relatively accurate correlation $R$, but with obvious absolute errors. Therefore, we simply develop one overall performance model to calibrate gem5~\cite{binkert2011gem5}. Specially, we directly use the ratio between ground-truth execution cycle numbers and the gem5 evaluation as the training label. 
The input features include all configuration parameters $C$ and selected event parameters $E$. Considering the branch prediction and long latency of memory access are critical for the performance, we select these key event parameters from $E$ as features: \{numCycles, idleCycles, branch\-Pred \-cond\-Pre\-dict\-ed, branch\-Pred \-cond\-In\-cor\-rect, icache \-over\-all\-Miss\-es, icache \-Read\-Req.\-mshr\-Miss\-es, dcache \-Read\-Req.\-misses, dcache \-Write\-Req.\-misses, dcache \-over\-all\-Miss\-es, dcache \-over\-all\-Mshr\-Miss\-es\}. 
(3) Finally, PANDA can naturally evaluate the energy consumption of a given workload, by combining its performance prediction with power prediction.

\subsection{Transferring to New Technology}


As the process technology node keeps shrinking, when power model is trained on labels from a certain technology node, it is often desirable to know the power of an unknown design in a new technology node. This task is obviously challenging. Designers often simply scale the power consumption from known technology to a new technology based on $P=CV^2$, but it is over-simplified. We propose a new ML method to predict the power consumption of \textcolor{black}{an} unknown configuration under a new unknown technology node. It naturally involves two steps. 1) Predict power of the unknown configuration with PANDA. The result corresponds to the technology node where PANDA is trained. We refer to it as \textcolor{black}{the} source node. 2) Based on the result, transfer the prediction to the new unknown technology node, named \textcolor{black}{the} target node. The insight to achieve the transferring is, when transferring to a new technology node, the impact on large designs i.e. BOOM CPU, and those very small designs is similar. It gives us an opportunity to predict the transformation of large designs using the ground-truth of small designs, which can be generated without too much cost.


We build an ML model for this transferring task. For each design, we use the ratio between power under target technology and source technology as the label. The features include (1) PANDA predicted power consumption under source node, (2) the ratio between target and source's scale and voltage. For example, when the source node is 28nm 0.8V, the target one is 40nm 1.1V, the feature is $40/28$ and $1.1/0.8$, (3) the directly-scaled power using $P=CV^2$. Take the previous 28nm to 40nm example, this scaled power is PANDA-predicted power at 28nm multiplied by $40/28 * (1.1/0.8)^2$. 
We collect ground-truth from small designs under multiple different technology nodes and train this transferring model only using small designs. Please notice that only one model is trained to perform transferring among multiple technology nodes. After training, this model can make transfer predictions between any two technology nodes. 



%% file: _txt/5_experiment.tex
\section{Experimental Results}

\subsection{Experiment Setup}

To evaluate our method, we generate a dataset by collecting RTL code and \textcolor{black}{performing} RTL simulation, with Chipyard~\cite{amid2020chipyard} v1.8.1. For \textcolor{black}{a} fair comparison with prior works~\cite{zhai2022mcpat}, we employed 15 similar RISC-V BOOM~\cite{zhao2020sonicboom} CPU configurations in Table~\ref{configtable}, named C1 to C15, ranging from small to large design sizes. Similar to prior works~\cite{bai2021boom, zhai2023microarchitecture}, we further divided these 15 configurations into five domains based on their \emph{DecodeWidth}, which is a key configuration parameter that affects multiple pipeline stages. The detailed configurations are given in Table~\ref{configtable}. For vector-based power simulation, we used eight workloads in the riscv-tests~\cite{URL:riscvtests} suite, including dhrystone, median, multiply, qsort, rsort, towers, spmv, and vvadd.

We performed RTL simulation at 1GHz with Synopsys VCS\textsuperscript{\textregistered}~\cite{vcs}. The logic synthesis and ground-truth power simulation are performed with Synopsys Design Compiler\textsuperscript{\textregistered}~\cite{design-compilier} and PrimePower~\cite{ptpx}, respectively. For the technology node, we used the TSMC 40nm standard cell library and the corresponding Memory Compiler. Furthermore, in our evaluation of cross-technology node prediction, we also adopt the TSMC 28nm and 65nm standard cell libraries.

\setcounter{figure}{4} 
\begin{figure*}[!b]
\centering
\hspace{-6mm}
\subfigure[McPAT]{
    \centering
    \includegraphics[height=0.16\textwidth]{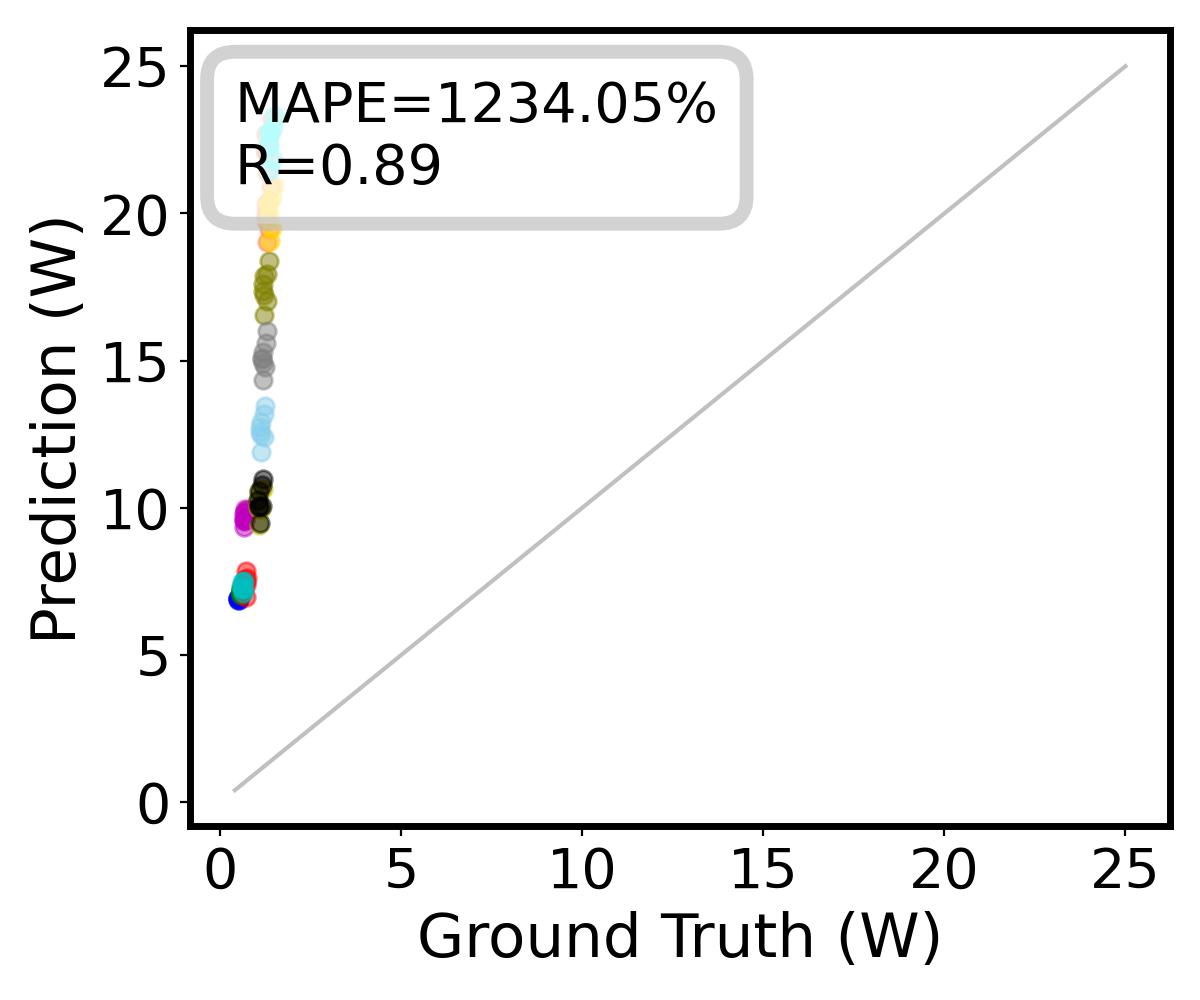}
    \label{unknown_0_mcpat}
}
\hspace{-3mm}
\subfigure[McPAT plus]{
    \centering
    \includegraphics[height=0.16\textwidth]{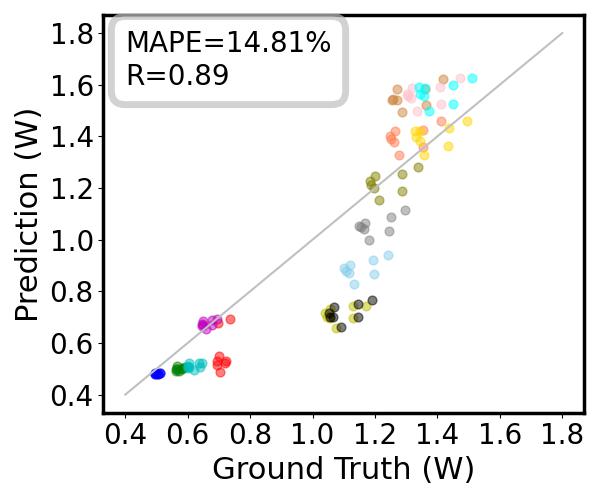}
    \label{unknown_0_mcpat_plus}
}
\hspace{-3mm}
\subfigure[McPAT-Calib]{
    \centering
    \includegraphics[height=0.16\textwidth]{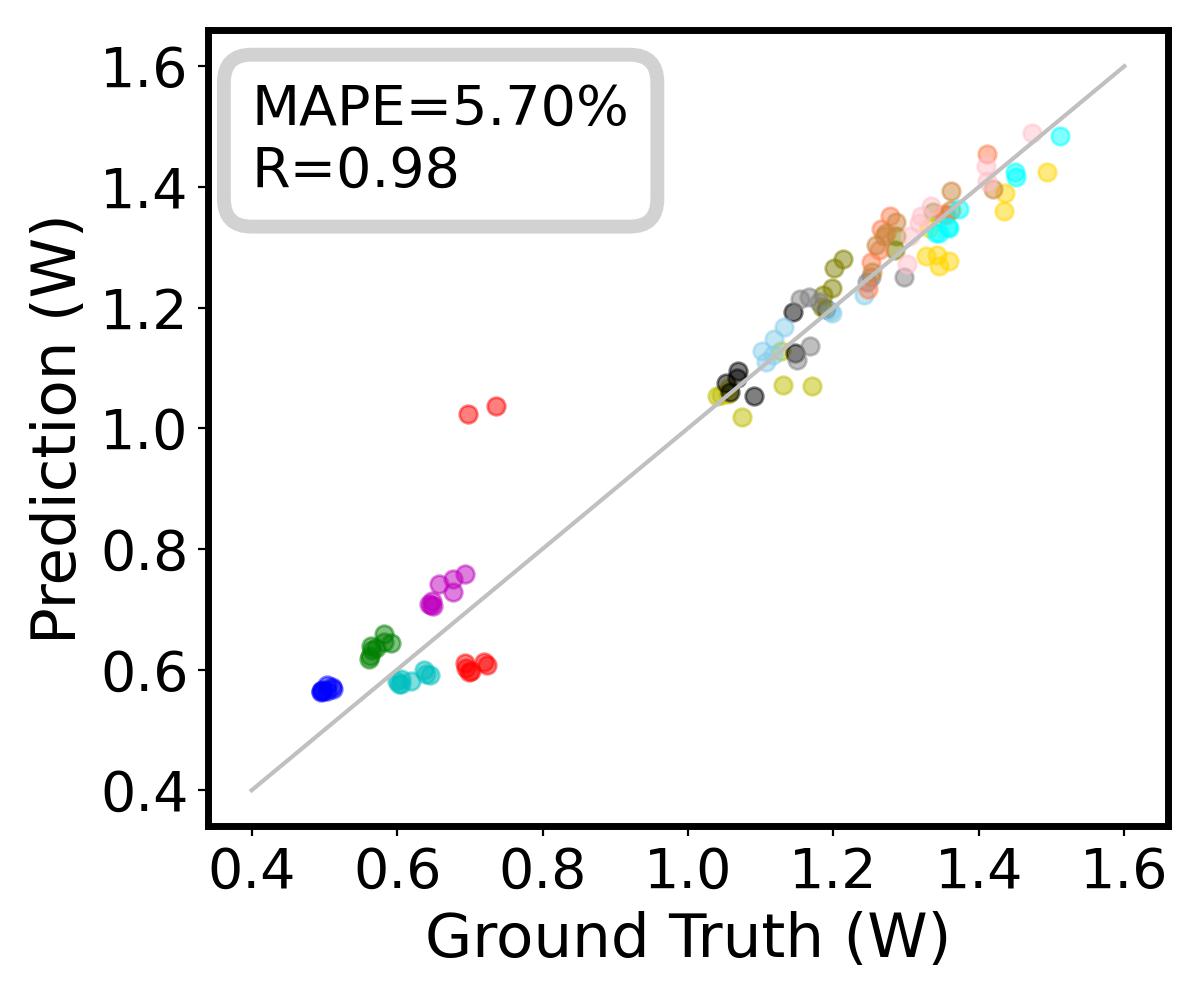}
    \label{unknown_0_base_claimed}
}
\hspace{-3mm}
\subfigure[Component-level Model]{
    \centering
    \includegraphics[height=0.16\textwidth]{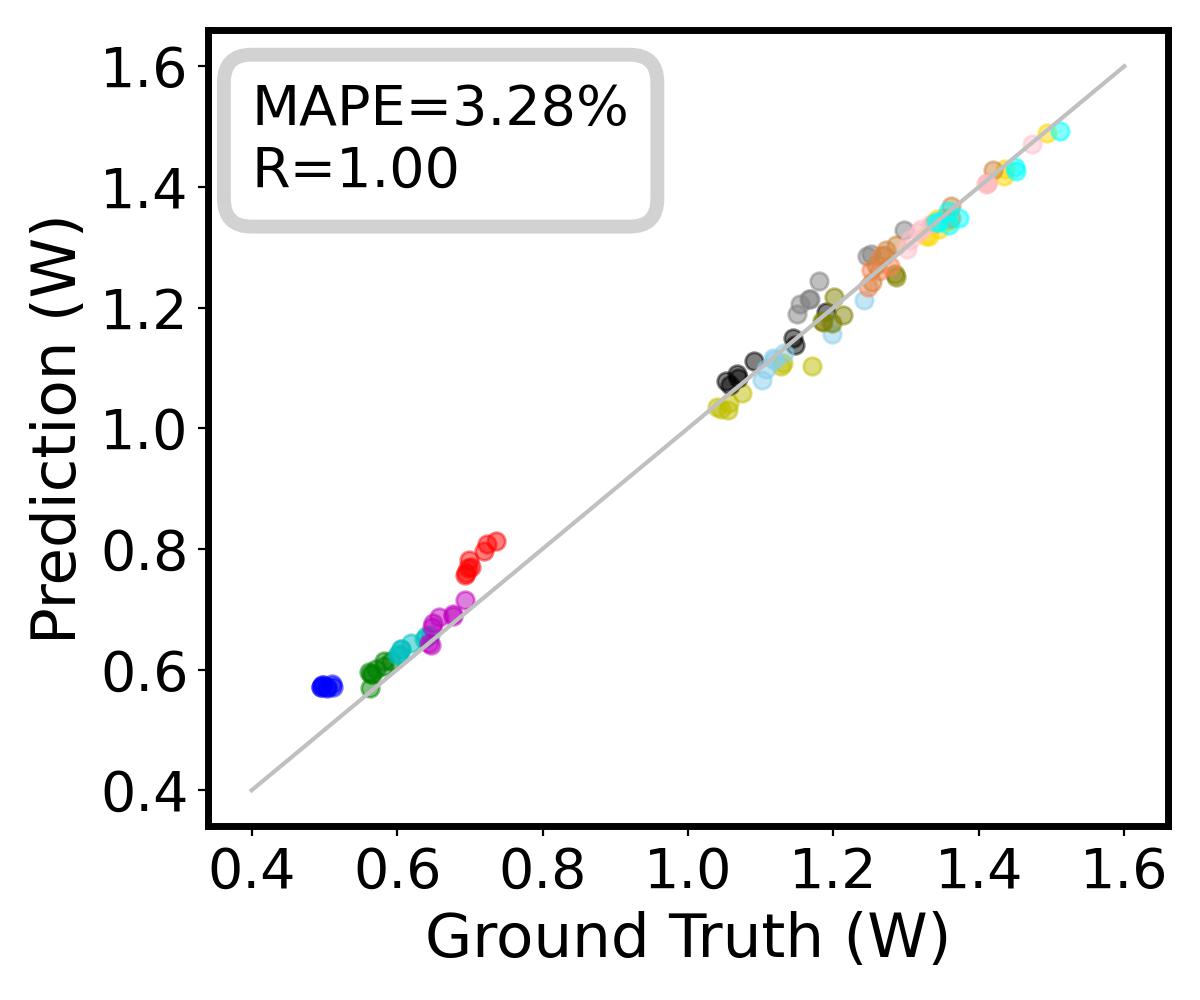}
    \label{unknown_0_comp}
}
\hspace{-3mm}
\subfigure[PANDA]{
    \centering
    \includegraphics[height=0.16\textwidth]{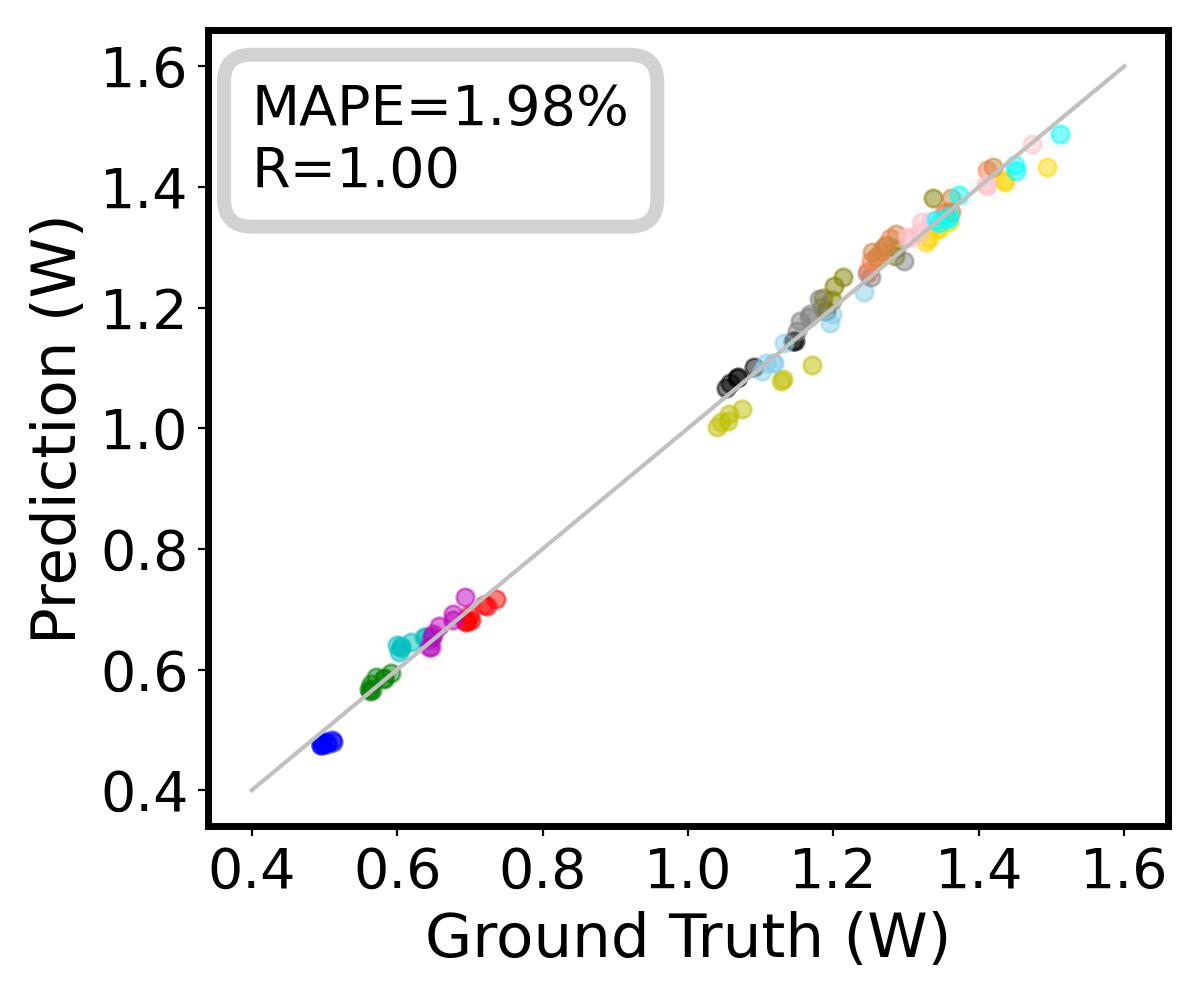}
    \label{unknown_0_our}
}
\vspace{-.1in}
\caption{Accuracy comparison for the known 14 (unknown 1) configuration scenario.}
\label{unknown_0}
\end{figure*}

\begin{figure*}[!b]
\centering
\vspace{-.2in}
\hspace{-6mm}
\subfigure[McPAT]{
    \centering
    \includegraphics[height=0.16\textwidth]{_fig/dynamic_mcpat.jpg}
    \label{unknown_d_mcpat}
}
\hspace{-3mm}
\subfigure[McPAT plus]{
    \centering
    \includegraphics[height=0.16\textwidth]{_fig/mcpat_plus_correct.jpg}
    \label{unknown_d_mcpat_plus}
}
\hspace{-3mm}
\subfigure[McPAT-Calib]{
    \centering
    \includegraphics[height=0.16\textwidth]{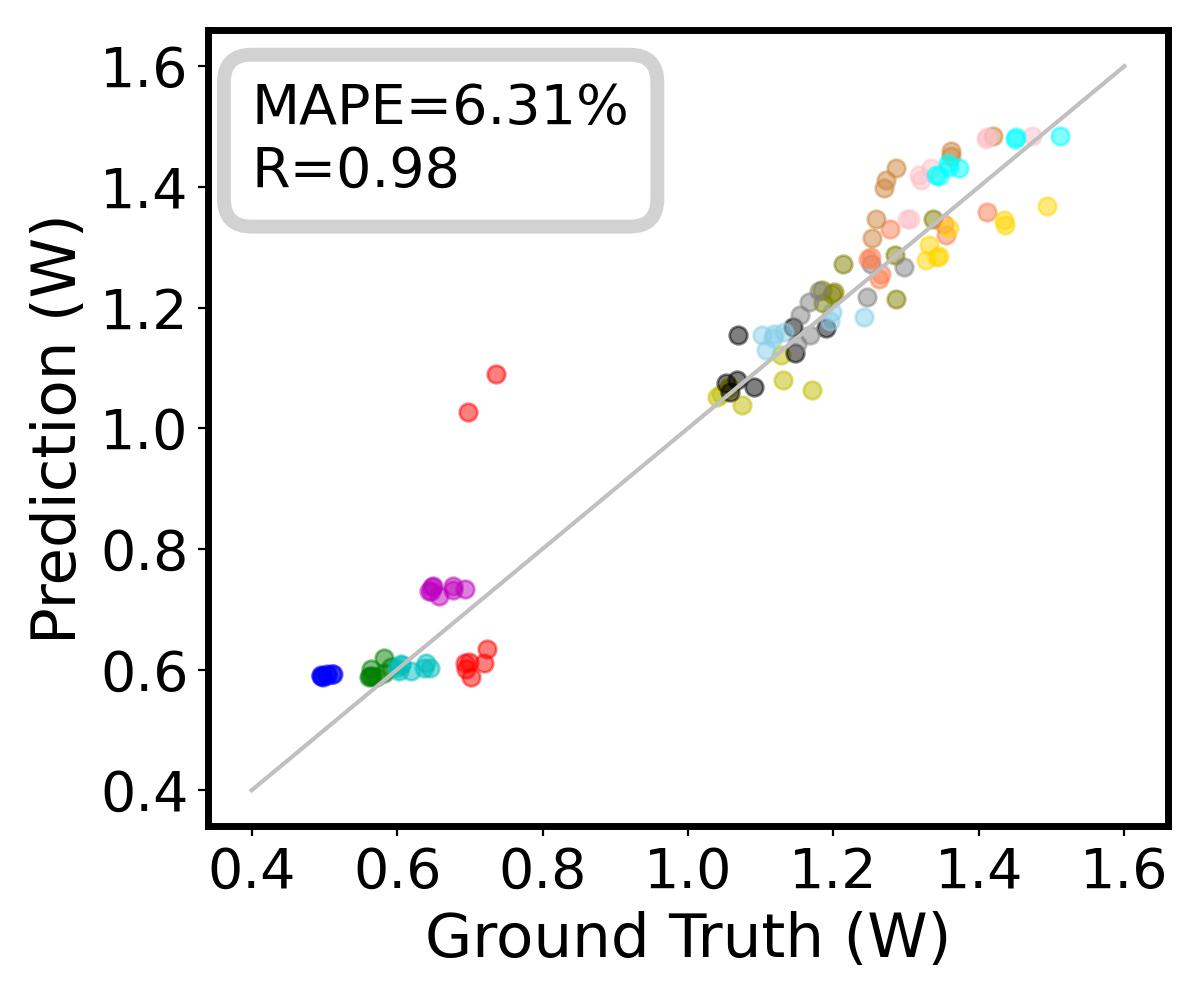}
    \label{unknown_d_base_claimed}
}
\hspace{-3mm}
\subfigure[Component-level Model]{
    \centering
    \includegraphics[height=0.16\textwidth]{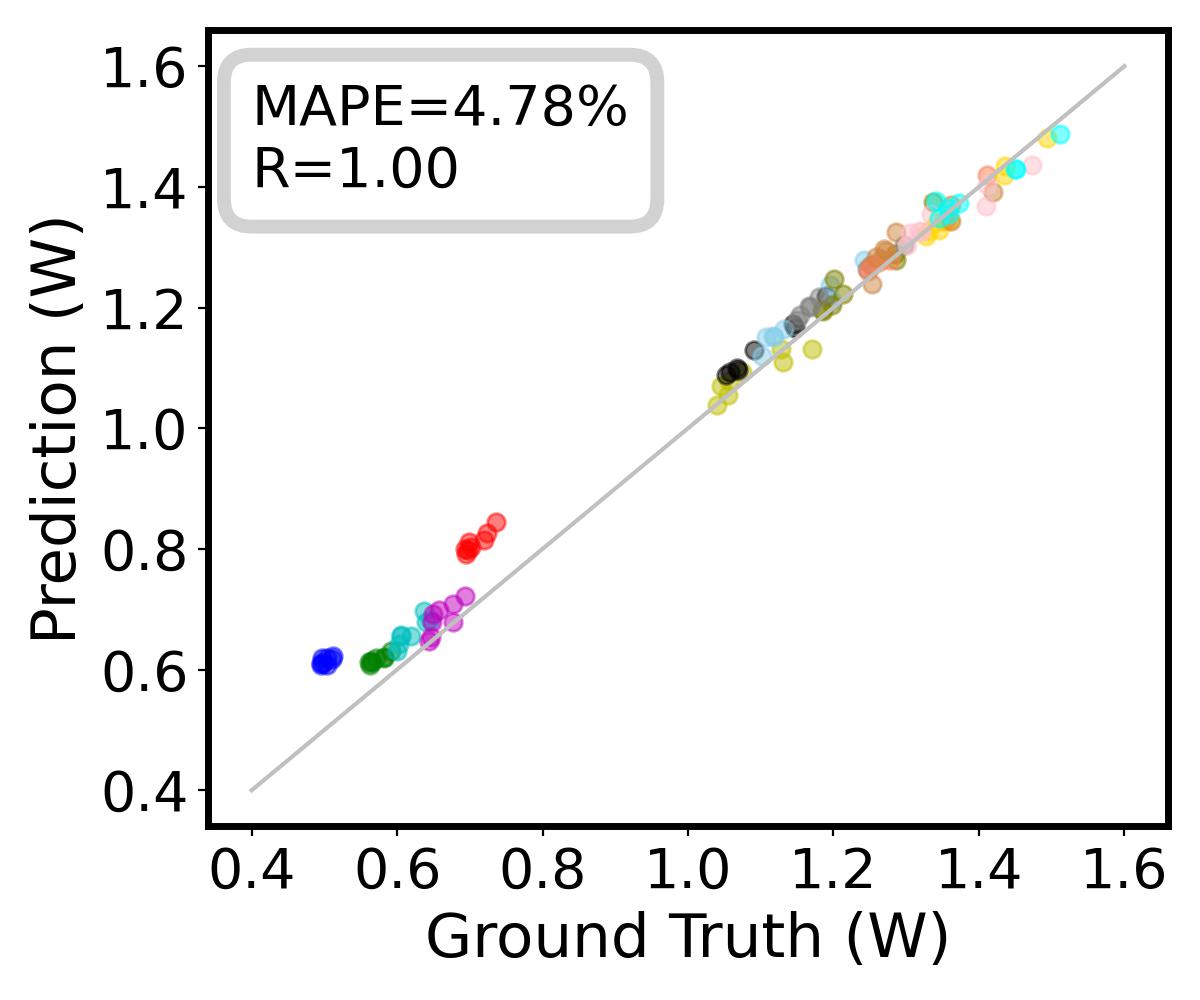}
    \label{unknown_d_comp}
}
\hspace{-3mm}
\subfigure[PANDA]{
    \centering
    \includegraphics[height=0.16\textwidth]{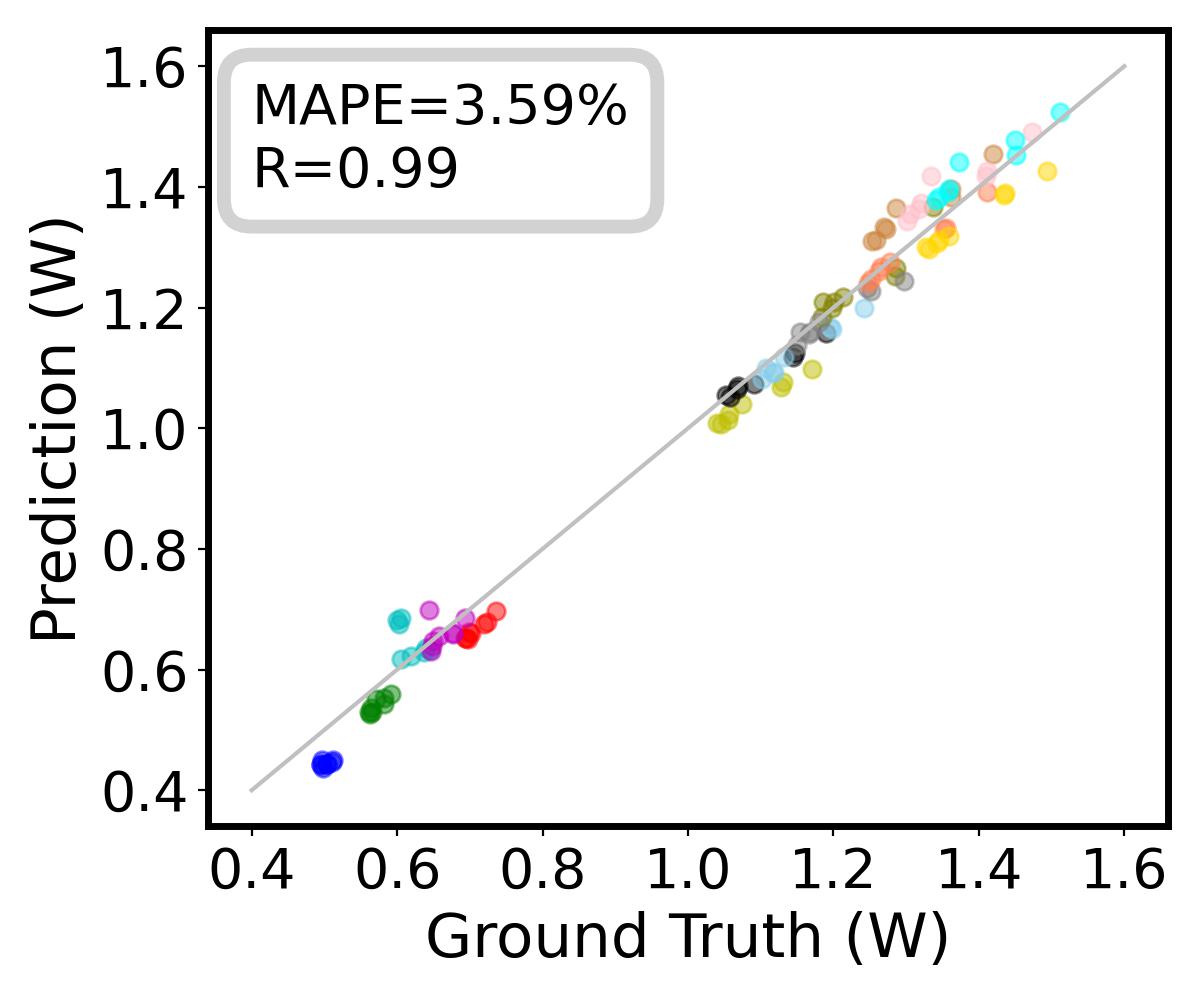}
    \label{unknown_d_our}
}
\vspace{-.1in}
\caption{Accuracy comparison for the unknown-design-domain scenario. } 
\label{unknown_d}
\end{figure*}

\subsection{Summary of Baseline Methods}

We compared PANDA with representative prior works, including (a) McPAT (+gem5~\cite{binkert2011gem5}~\cite{roelke2017risc5})~\cite{li2009mcpat}, a typical analytical model, (b) PowerTrain~\cite{lee2015powertrain}, a widely-adopted lightweight ML-based model, and (c) McPAT-Calib~\cite{zhai2022mcpat}, which is the state-of-the-art ML-based model. All of these three works \textcolor{black}{rely} on McPAT as part of the power model. Additionally, (d) TCAD'17~\cite{walker2016accurate}, a representative on-chip power meter design method based on performance counters, is included as a baseline. On-chip power meters are not originally designed for architecture-level power modeling, we include \textcolor{black}{them} for the completeness of our experiment. For this task, the performance counters are replaced with the event parameters generated by gem5~\cite{binkert2011gem5}.

For a recent work ASPDAC'23~\cite{zhai2023microarchitecture}, it mainly proposed a transfer learning algorithm to predict the power of configurations in \textcolor{black}{the} unknown domain. But we emphasize that such transfer learning still requires sampling in the unknown domain, which necessitates the RTL implementation and subsequent synthesis of the sampled configurations. It is quite costly, since although the Chipyard can generate RTL code automatically, it is not a always the case in the industry, where implementing RTL for a new CPU configuration may require significant engineering effort. What's more, the core power prediction part of ASPDAC'23 is similar to McPAT-Calib~\cite{zhai2022mcpat}, by adopting multi-layer perceptron as its ML model. Therefore, we do not present it separately.

Besides using prior works as baselines, we also include two extra baselines. (e) To further demonstrate the potential of McPAT without inaccuracies introduced by its internal coefficients, \textcolor{black}{an} ideal scaling factor is derived based on ground-truth power of all configurations. It scales the power prediction of McPAT towards the ground-truth. This baseline with superior accuracy than McPAT is named McPAT-plus. (f) We also build a much weaker variant of PANDA with limited architecture-level knowledge, named the Component-level model, as shown in Fig.~\ref{abstract2}. It builds ML models for each component with related configuration parameters and event parameters in Table~\ref{tbl:config_event} as features. But it does not adopt the resource functions in PANDA.

\setcounter{figure}{3} 
\begin{figure}[!t]
\centering
\includegraphics[width=0.48\textwidth]{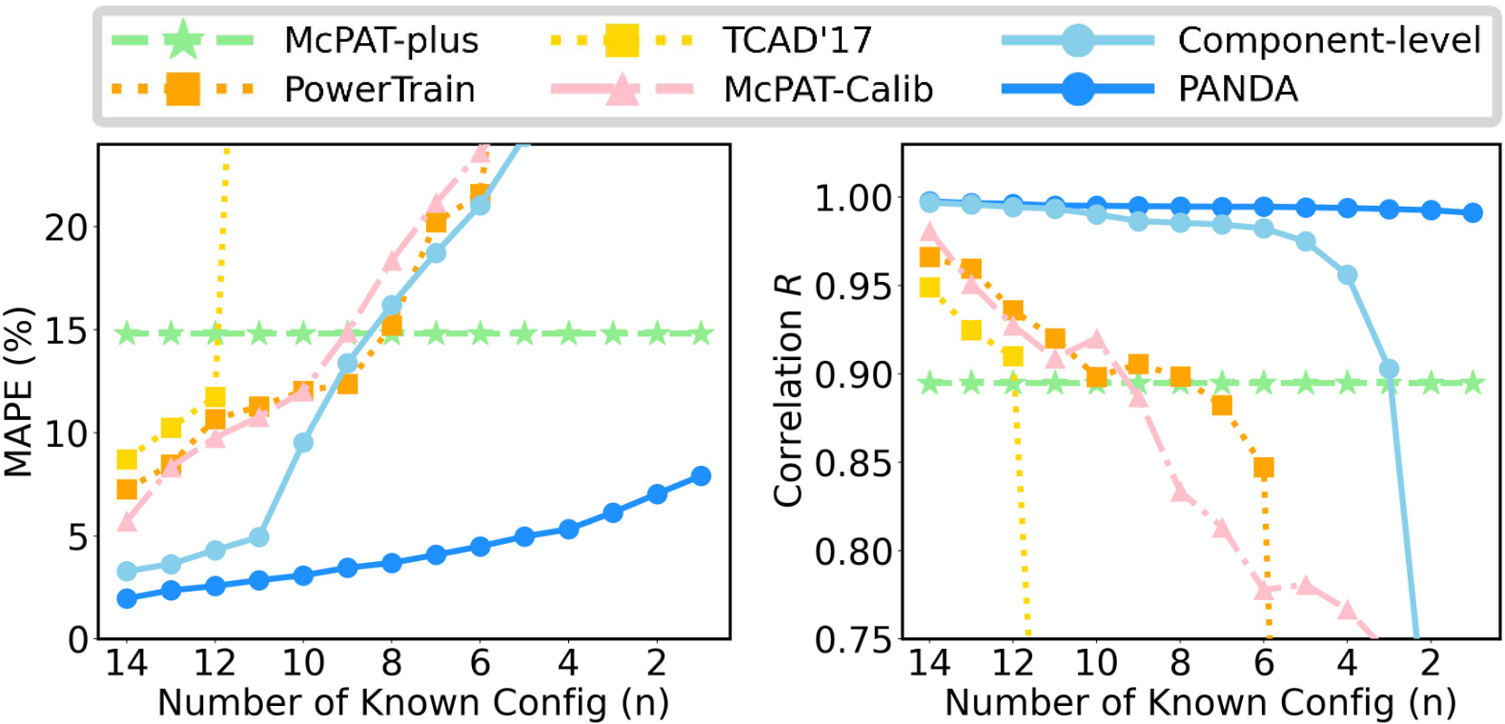}
\caption{The MAPE and $R$ of different models under different \textcolor{black}{number} of unknown configurations. The McPAT-plus is a correctly calibrated version of McPAT, directly scaled with ground-truth. PANDA's superior accuracy over baselines increases as the training data decrease. } 
\vspace{-.1in}
\label{diff_unknown}
\end{figure}

We conduct multiple experiments with different \textcolor{black}{amount} of training data. When the number of known configurations as training data is $n$ ($n \in [1, 14]$), the $15-n$ unknown configurations will be testing data\footnote{For example, when $n$ is 13, there are 13 known configurations for training and 2 testing configurations for testing. We sequentially traverse testing sets with neighboring configurations (C1,C2), ..., (C14,C15), and (C15,C1). Altogether 15 models are trained \& evaluated, corresponding to each testing set. These predictions on each testing configuration are averaged as \textcolor{black}{the} final prediction.}. The experiment setup and actual data used for training and testing are strictly the same for all methods during comparison. 
We evaluate performance with the mean absolute percentage error (MAPE)\footnote{$\text{MAPE}= 1/n * \sum_{k=1}^{n} {|y_k-\hat{y_k}|/y_k}$, $y_k$ is label and $\hat{y_k}$ is prediction.} and correlation coefficient $R$ averaged over all testing configurations.


\setcounter{figure}{6} 
\begin{figure*}[!t]
\centering
\vspace{-.2in}
\hspace{-6mm}
\subfigure[McPAT]{
    \centering
    \includegraphics[height=0.16\textwidth]{_fig/dynamic_mcpat.jpg}
    \label{unknown_10_mcpat}
}
\hspace{-3mm}
\subfigure[McPAT-plus]{
    \centering
    \includegraphics[height=0.16\textwidth]{_fig/mcpat_plus_correct.jpg}
    \label{unknown_10_mcpat_plus}
}
\hspace{-3mm}
\subfigure[McPAT-Calib]{
    \centering
    \includegraphics[height=0.16\textwidth]{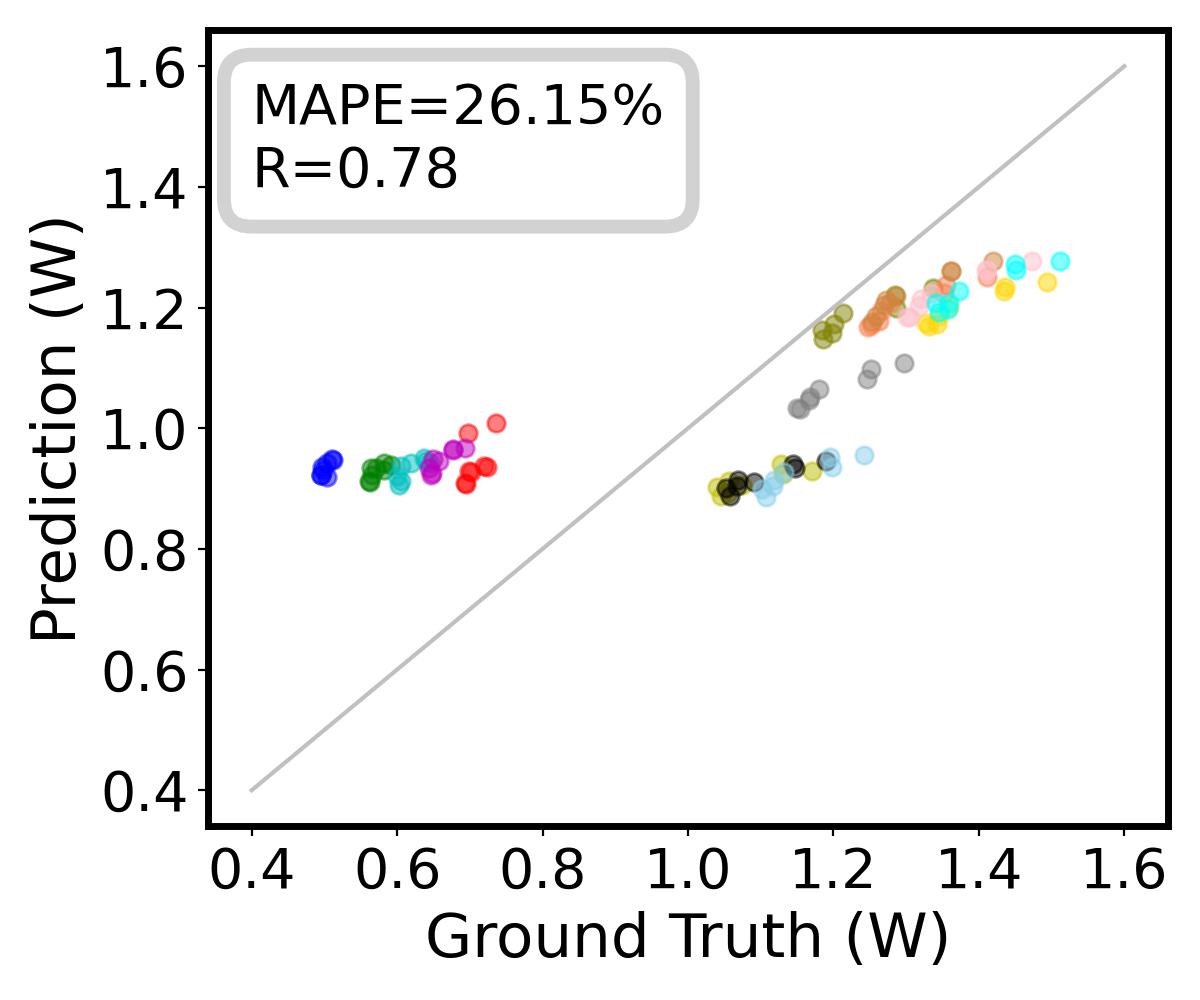}
    \label{unknown_10_base_claimed}
}
\hspace{-3mm}
\subfigure[Component-level Model]{
    \centering
    \includegraphics[height=0.16\textwidth]{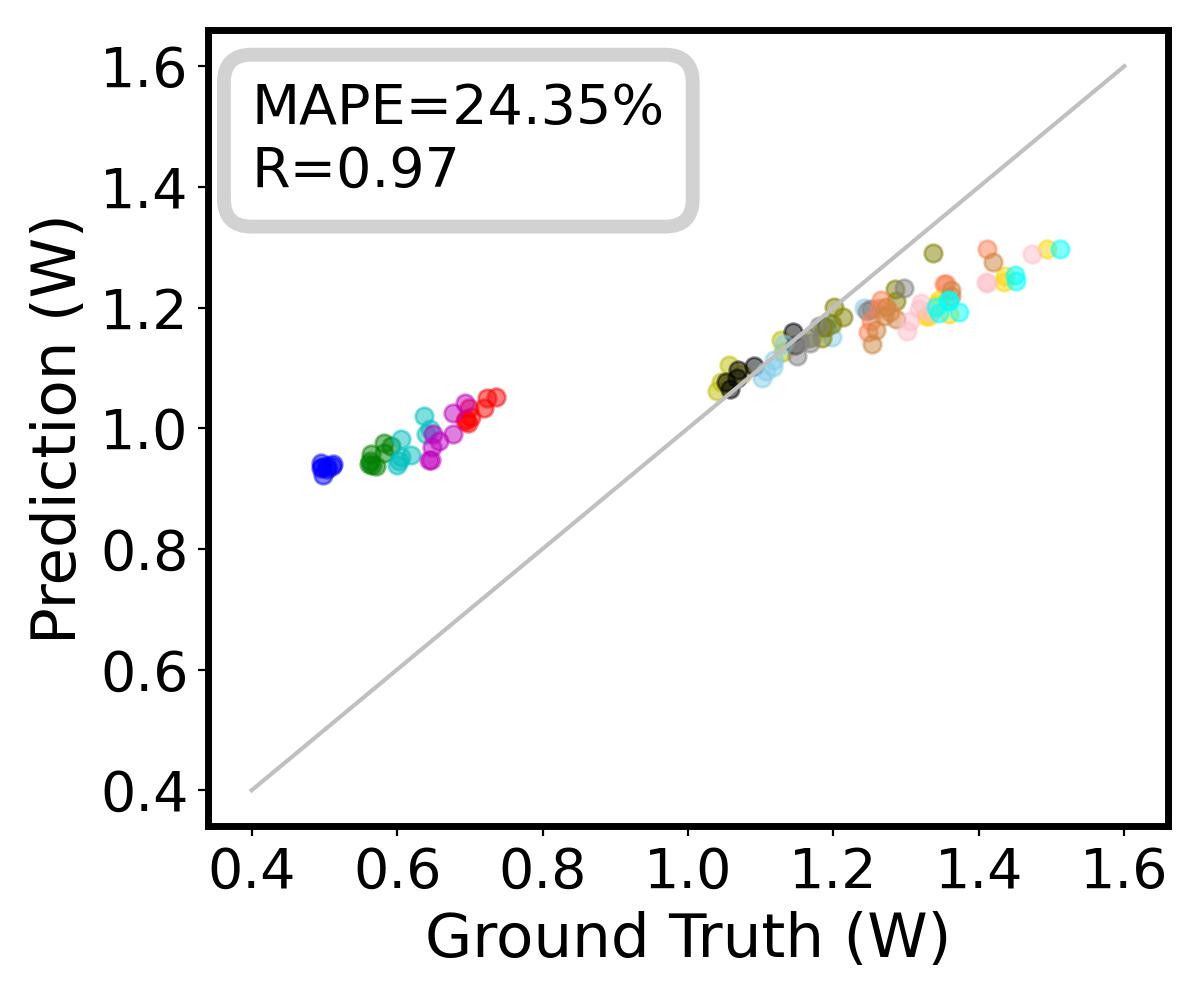}
    \label{unknown_10_comp}
}
\hspace{-3mm}
\subfigure[PANDA]{
    \centering
    \includegraphics[height=0.16\textwidth]{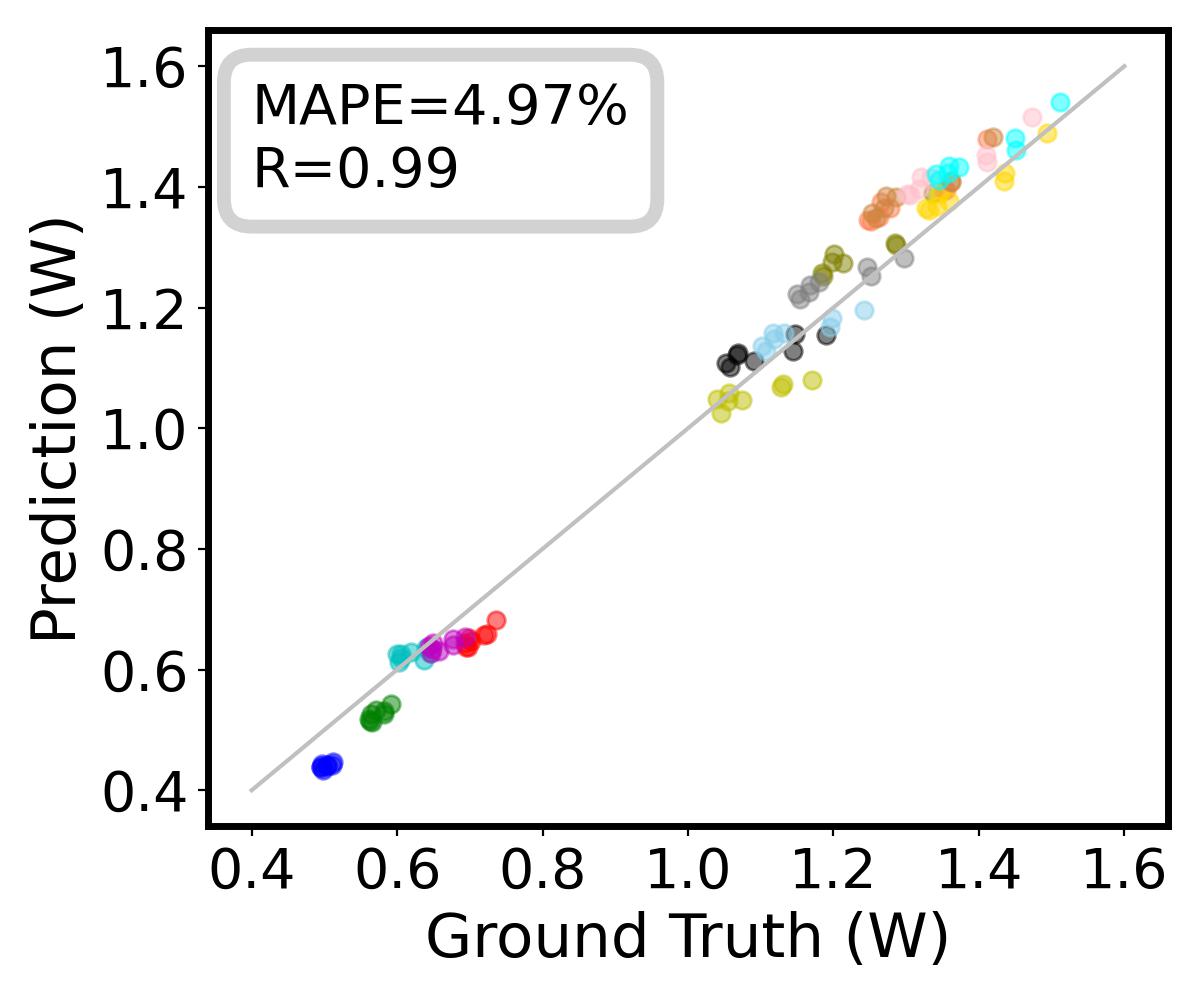}
    \label{unknown_10_our}
}
\vspace{-.1in}
\caption{Accuracy comparison for the known 5 (unknown 10) configuration scenario.}
\label{unknown_10}
\end{figure*}

\begin{figure*}[!t]
\centering
\vspace{-.2in}
\hspace{-6mm}
\subfigure[McPAT]{
    \centering
    \includegraphics[height=0.16\textwidth]{_fig/dynamic_mcpat.jpg}
    \label{unknown_14_mcpat}
}
\hspace{-3mm}
\subfigure[McPAT plus]{
    \centering
    \includegraphics[height=0.16\textwidth]{_fig/mcpat_plus_correct.jpg}
    \label{unknown_14_mcpat_plus}
}
\hspace{-3mm}
\subfigure[McPAT-Calib]{
    \centering
    \includegraphics[height=0.16\textwidth]{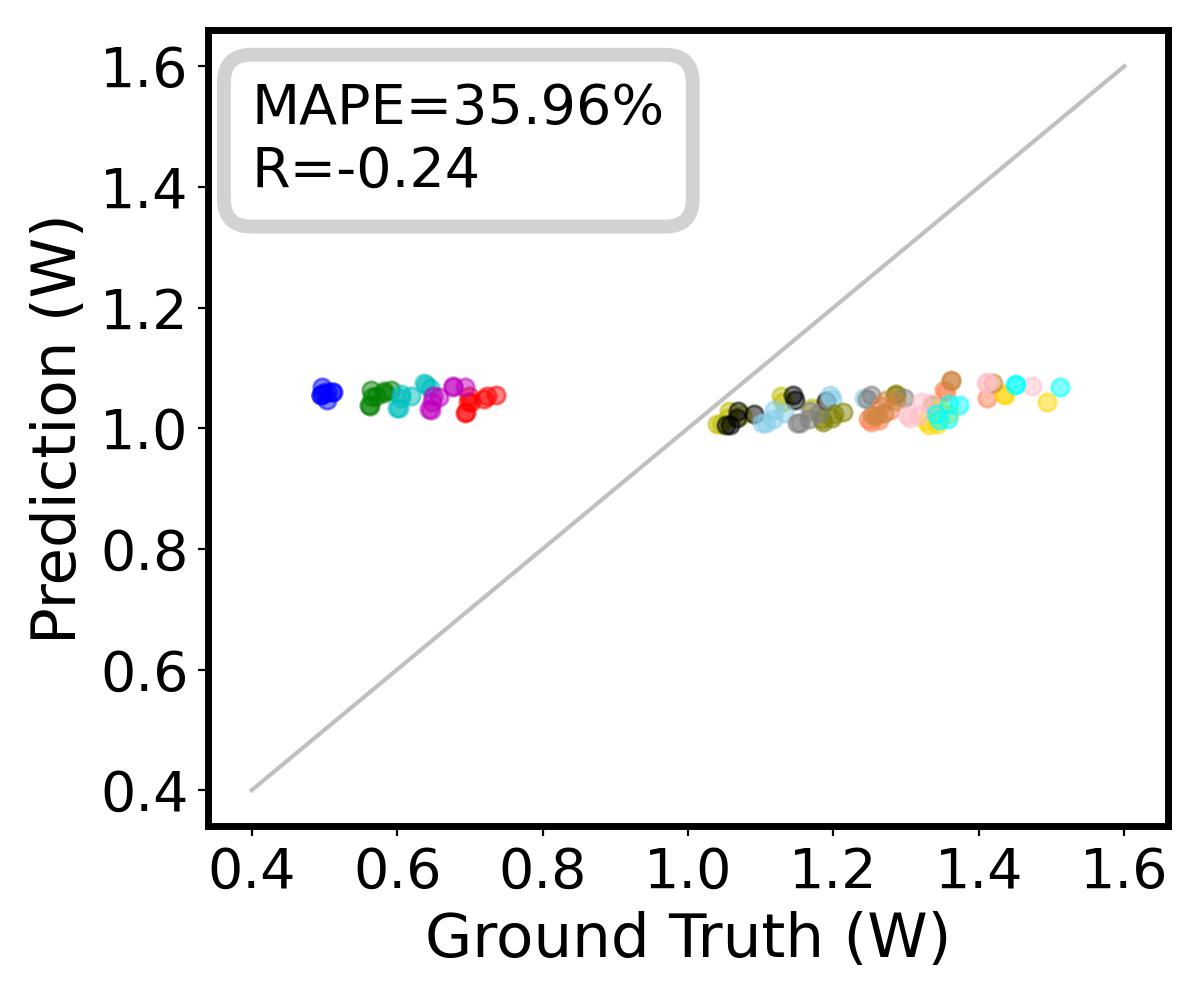}
    \label{unknown_14_base_claimed}
}
\hspace{-3mm}
\subfigure[Component-level Model]{
    \centering
    \includegraphics[height=0.16\textwidth]{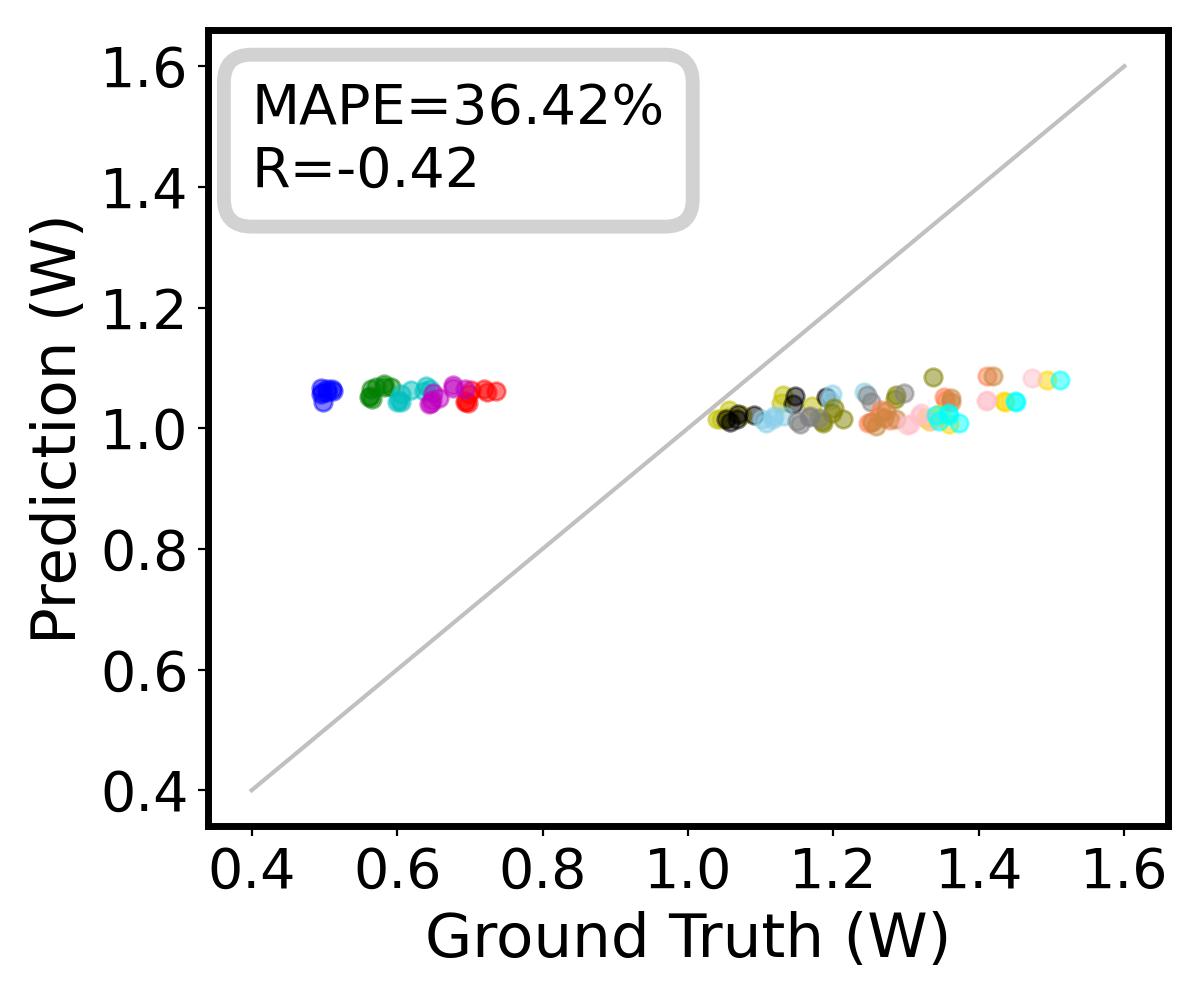}
    \label{unknown_14_comp}
}
\hspace{-3mm}
\subfigure[PANDA]{
    \centering
    \includegraphics[height=0.16\textwidth]{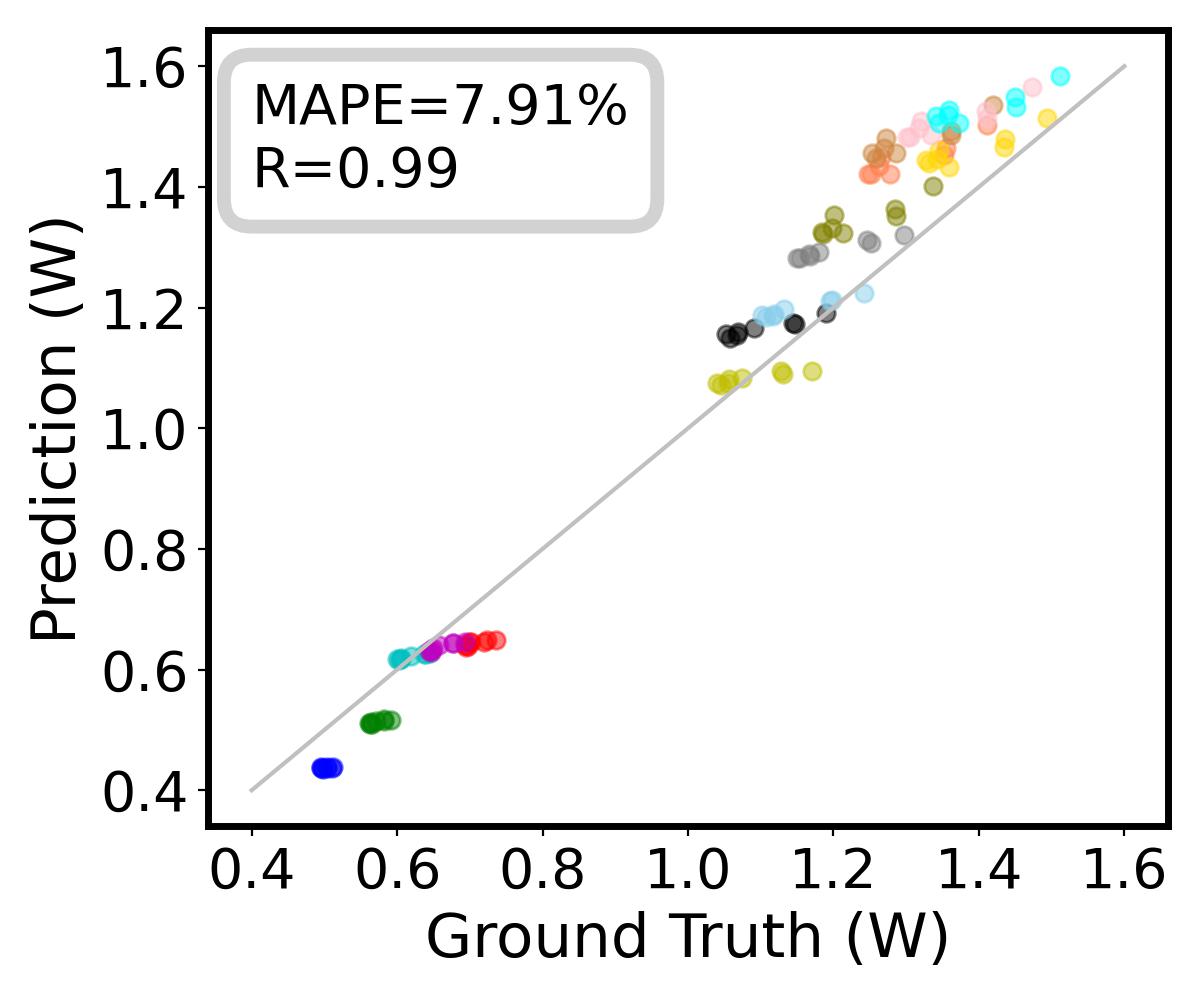}
    \label{unknown_14_our}
}
\vspace{-.1in}
\caption{Accuracy comparison for the known 1 (unknown 14) configuration scenario.}
\label{unknown_14}
\end{figure*}

\subsection{Power Prediction Results}

Fig.~\ref{diff_unknown} shows the accuracy of PANDA and our baseline models when they are trained with different number of known configurations. PANDA consistently achieves the lowest MAPE and highest $R$. The superior performance of PANDA over all ML-based baselines is increasingly obvious as the number of known \textcolor{black}{configurations} (i.e., training data amount) in \textcolor{black}{the} x-axis decreases. This trend validates PANDA's excellent accuracy given very limited training data.

Here we try to analyze the reasons behind the performance gap between PANDA and representative ML solutions McPAT-Calib~\cite{zhai2022mcpat} and PowerTrain~\cite{lee2015powertrain} in Fig.~\ref{diff_unknown}. By integrating architecture-level knowledge into its resource function, PANDA can efficiently capture how the configuration parameters modification will affect the power based on very limited training data. In comparison, in previous ML solutions, too much knowledge needs to be learned when training a single ML model from scratch. Also, since they rely on McPAT, they may be limited by McPAT's poor accuracy when approaching a higher accuracy. 
Another ML baseline TCAD'17~\cite{walker2016accurate} naturally performs poorly since it is designed for on-chip power meter development instead of this task.

As for analytical model baseline McPAT-plus, as a correctly-scaled version of McPAT, its accuracy remains unchanged in Fig.~\ref{diff_unknown} regardless of training data amount. Without ML model, its overall accuracy is limited. 
By unifying both analytical and ML techniques, PANDA outperforms it significantly even when there is only one known configuration ($n=1$)\footnote{For $n=1$, as a special case, the bias in a few resource function has to be derived based on designers observation. But it is strictly fitted with training data when $n\geq2$.} for training. For the original McPAT, while the $R$ is the same as McPAT-plus, the absolute error MAPE is higher than 1000\%, so it is not presented in the figure.




Finally, the component-level model, as a weaker variant of PANDA with limited architecture knowledge, provides a great decomposition of PANDA's high performance. Its comparison with PANDA can be viewed as a simple ablation study. By training ML models at component level, it maintains a reasonable accuracy when the number of known configurations $n>10$ and outperforms other ML baselines. But PANDA soon outperforms it significantly when training data further decreases. Such a gap shows the contribution of resource functions with architecture knowledge.


In Fig.~\ref{unknown_0}, \ref{unknown_10}, \ref{unknown_14},  we further visualize the detailed prediction results when the number of known configurations ($n$) for training is 14, 5, and 1, respectively. Additionally, in Fig.~\ref{unknown_d} we introduce a new scenario called the `unknown domain', where each time there are 4 domains as known training sets (with 12 configurations) and 1 domain as the testing set (with 3 configurations). The testing set will traverse all 5 domains to generate predictions on all 15 configurations. 




\subsection{Prediction of Other Design Qualities}

\begin{table}[!t]
      \centering
      \renewcommand{\arraystretch}{1.1}
      \resizebox{0.49\textwidth}{!}{
        \begin{tabular}{ |c||c|c|c||c|c| } 
        \hline
        \multirow{2}{*}{Design Quality} & \multicolumn{3}{c||}{Baseline} & \multicolumn{2}{c|}{PANDA} \\ 
        \cline{2-6}
        &  Baseline Method  &  MAPE(\%) & $R$ &  MAPE(\%) & $R$ \\
        \hline
        \hline
        Area & McPAT & 416.56 & 0.98 & 2.92 & 0.99 \\
    \hline
    Performance & gem5 & 26.79 & 0.98 & 6.69 & 0.98 \\
    \hline
    Energy & gem5 + McPAT-Calib & 31.87 & 0.97 & 9.51 & 0.98 \\
\hline
        \end{tabular}
        }
        \caption{The comparison of Area, Performance, and Energy prediction between baseline and PANDA}
        \label{other_quality}
        \vspace{-.1in}
\end{table}


Besides power prediction, Table~\ref{other_quality} shows PANDA's prediction accuracy on area, performance, and energy. Since the majority of prior power models do not further evaluate these design qualities, in this part, we use McPAT as the baseline for area prediction, and gem5 as the baseline for performance prediction. For the energy baseline, we collect the performance predicted by gem5 and power predicted by McPAT-Calib to compute the energy. Table~\ref{other_quality} also adopts the `unknown-domain' scenario, where each time 1 unknown domain is used as \textcolor{black}{the testing} set. The baseline methods achieve good correlation in all 15 configurations but with a huge absolute error value. In comparison, PANDA achieves even higher correlation and keeps the error within 10\%.

\subsection{Cross-Technology Prediction}
\label{sec:cross_tech}

We evaluate the cross-technology prediction accuracy of PANDA with three different technology nodes, TSMC 28nm 0.8V, TSMC 40nm 1.1V, and TSMC 65nm 1.2V. Our proposed transfer model is trained based on about 20 small designs implemented with all three technologies. On average, these small designs consist of only thousands of gates, in contrast with 0.3 million gates in the smallest BOOM configuration C1. For any pair of source and target technology nodes, the transfer model will predict the power of an unknown design configuration at the target technology node, based on PANDA power model's prediction at the source node. 


We calculate 1) the original prediction of source technology based on PANDA's power model; 2) the directly-scaled prediction towards the target technology based on $CV^2$; 3) the transferred prediction with our proposed transferring model. We compare these predictions with ground-truth labels at the target technology node, evaluating the MAPE. \textcolor{black}{Their accuracies are shown in Table ~\ref{cross_tech_node}.} The average MAPE of scaled prediction and model prediction are 22.16\% and 10.40\%, respectively, showing that PANDA outperforms analytically scaling.

\begin{table}[!t]
      \centering
      \renewcommand{\arraystretch}{1.1}
      \resizebox{0.48\textwidth}{!}{
        \begin{tabular}{ |c|c||c|c|c| } 
        \hline
        Source & Target & MAPE-Original(\%) & MAPE-Scaled(\%) & MAPE-PANDA(\%) \\  
        \hline
        \hline
        28 nm & 40 nm & 51.51 & 30.98 & 14.83 \\
\hline
28 nm & 65 nm & 73.12 & 40.42 & 10.16 \\
\hline
40 nm & 28 nm & 115.98 & 20.03 & 6.24 \\
\hline
40 nm & 65 nm & 43.39 & 10.07 & 11.66 \\
\hline
65 nm & 28 nm & 289.02 & 25.52 & 5.28 \\
\hline
65 nm & 40 nm & 83.94 & 5.91 & 14.24 \\
\hline
\hline
\multicolumn{2}{|c||}{Average} & 109.49 & 22.16 & 10.40 \\
\hline
        \end{tabular}
        }
        \vspace{.08in}
        \caption{Cross-technology prediction by 1) prediction at source technology, 2) directly-scaled prediction towards target technology, 3) prediction transferred to target technology by PANDA.}
        \label{cross_tech_node}
        \vspace{-.1in}
\end{table}

%% file: _txt/6_discussion.tex
\section{Discussion}

To further demonstrate the advantages of PANDA, we present two application scenarios as case studies. 
Finally, we present an analysis to \textcolor{black}{verify} the correctness of the design of PANDA.


\subsection{Case Study 1: Power Prediction for Special Configurations}

For existing 15 configurations in the experiment, from the smallest C1 to the largest C15, all component parameters monotonically increase with a similar trend. As a result, although the total power increases from C1 to C15, the power percentage of each component does not vary much. However, in practice, this is not always the case, as realistic configurations may have different percentages of power consumption for some key components. To study this scenario, we design two special cases SP1 and SP2, as already shown in Table~\ref{configtable}. SP1 is with a large BP but small other components, while SP2 is with a small D-Cache and I-Cache but large other components.
For such special-case designs, prior ML methods may perform poorly due to a lack of awareness of CPU hierarchy, which makes it difficult to capture how each component contributes to the total power consumption. In comparison, PANDA can well handle any configuration parameter combinations by modeling each component separately.





\begin{figure}[!t]
\centering
\vspace{-.16in}
\hspace{-6mm}
\subfigure[McPAT-Calib]{
    \centering
    \includegraphics[height=0.18\textwidth]{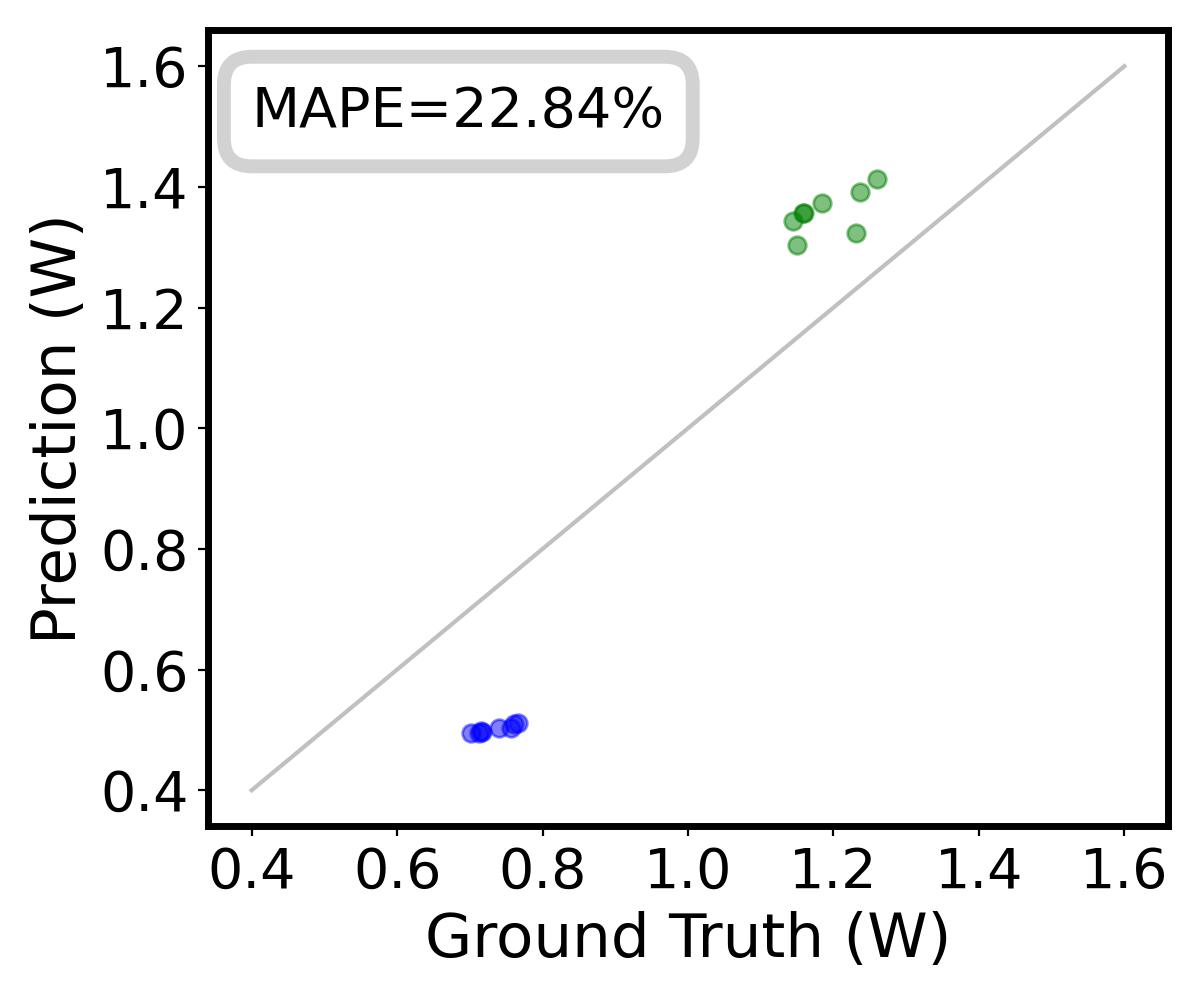}
    \label{extreme_case_base_claimed}
}
\hspace{-5mm}
\subfigure[PANDA]{
    \centering
    \includegraphics[height=0.18\textwidth]{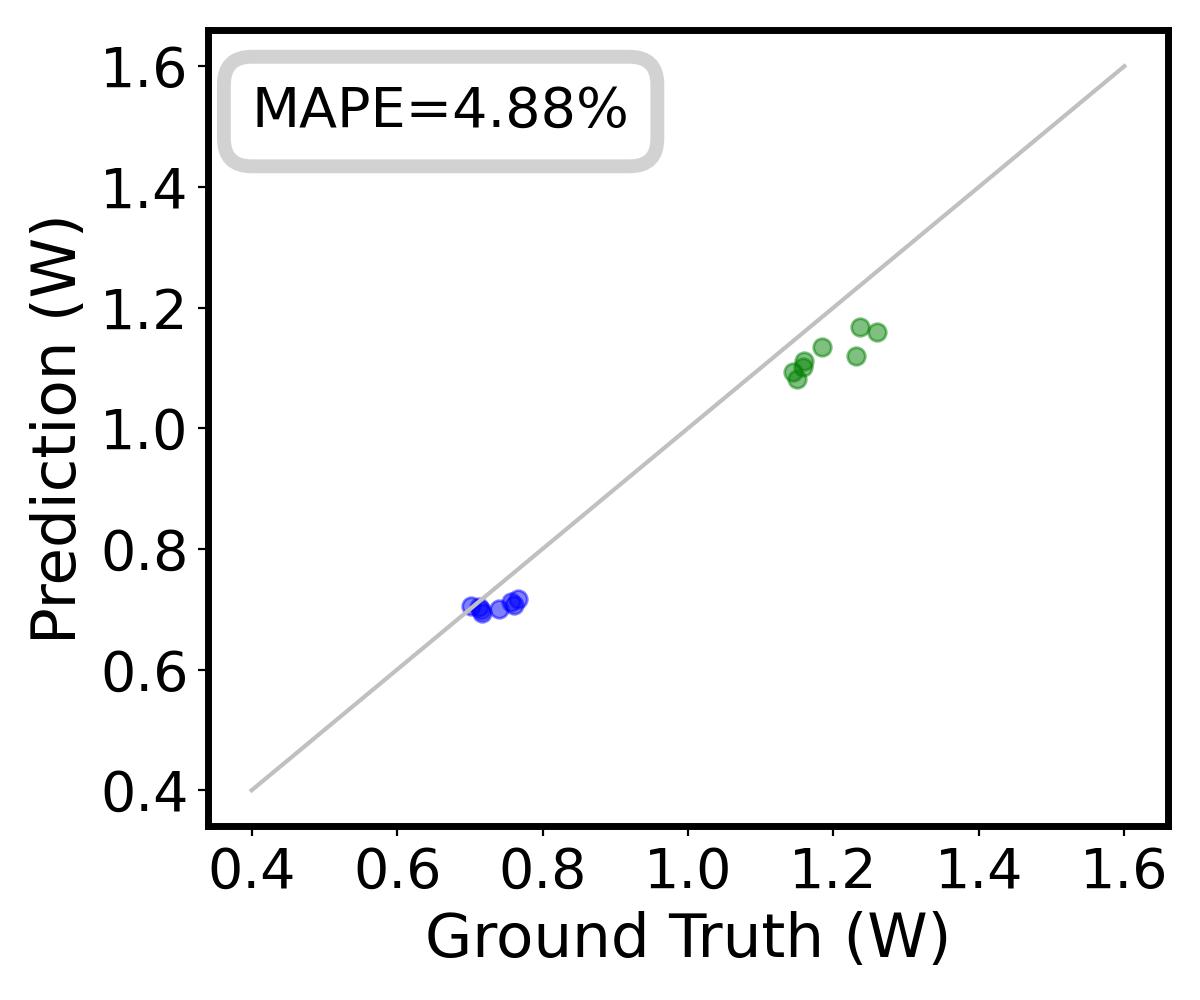}
    \label{extreme_case_our}
}
\vspace{-.1in}
\caption{The prediction of two special \textcolor{black}{cases} with different patterns in configuration parameters from the training set (C1-C15), blue means a CPU with \textcolor{black}{a} large Frontend but small other components, green means a CPU with \textcolor{black}{a} small I-Cache and D-Cache but large other components.}
\label{extreme_case}
\vspace{-.13in}
\end{figure}

Fig.~\ref{extreme_case} compares the predictions on SP1 and SP2 by McPAT-Calib and PANDA trained with C1 to C15, showing the MAPE\footnote{Since there are only two configurations, correlation $R$ is not a good metric.}. Despite all 15 configurations are used for training, McPAT-Calib is very inaccurate with MAPE=22.8\%. Compared with McPAT-Calib's average MAPE=5.7\% when trained with 14 known configurations in Fig.~\ref{diff_unknown}, this significantly lower accuracy indicates the challenges of special cases. 
In comparison, PANDA prediction remains accurate with MAPE$<$5\% for these special cases. It implies that PANDA can maintain a reasonably high accuracy on almost any new \textcolor{black}{configuration}. 


\subsection{Case Study 2: Design Space Exploration with PANDA}

\begin{table}[!b]
        \vspace{-.1in}
      \centering
      \renewcommand{\arraystretch}{1.1}
      \resizebox{0.49\textwidth}{!}{
        \begin{tabular}{ |c||c|c|c| } 
        \hline
        Design Name & Config Parameter (same order as TABLE~\ref{configtable}) & Power & Performance \\  
        \hline
        \hline
        Engineer Design & 4, 3, 24, 96, 96, 96, 24, 16, 2, 4, 2, 8, 8, 2 & 0.79 & 2.08\\
        \hline
        DSE Design & 4, 4, 32, 128, 96, 96, 32, 16, 2, 4, 2, 8, 4, 2 & 0.80 & 2.30\\
        \hline
        \end{tabular}
        }
        \caption{The comparison of design selected by engineer and design selected by DSE, the parameters are in the same order as TABLE~\ref{configtable}. Power unit is (W) and performance is normalized by C1's performance.}
        \label{dse}
\end{table}


Design space exploration (DSE) is an important task of CPU architects. For example, given a power constraint, architects explore which configuration achieves the highest performance. However, generating ground-truth power and performance of each design configuration requires timing-consuming design implementation. PANDA, as an accurate architecture-level power and performance model, naturally supports efficient DSE.

We assume a DSE task where configurations C1-C15 are known and PANDA is already trained by them. Since none of these configurations has \textcolor{black}{the} power value in the range of 0.7 to 1W, we let PANDA explore this unknown region. The specific task is \textcolor{black}{to} select a configuration with maximum performance, \textcolor{black}{with} power consumption less or equal \textcolor{black}{to} 0.8W. Here we compare the performance of PANDA and human engineers, who manually set configuration parameters using the ground-truth power of C1-C15 as a reference.

Then we start DSE by efficiently \textcolor{black}{predicting} almost all reasonable configuration parameter combinations in the design space with PANDA, then select high-performance configurations that \textcolor{black}{satisfy} the power constraint. We tolerate configurations with predicted power slightly higher than the constraint. This process is fast due to the efficiency of PANDA. We then start to implement selected configurations based on \textcolor{black}{the} rank of the predicted performance values. This process stops when one configuration's implementation proves to satisfy the power constraint. Due to the accuracy of PANDA, we reach the power constraint after very few trials. Human engineers go through a similar validation process, they craft and implement new configurations until power constraint is met. 


The ground-truth power and performance of PANDA-explored configuration \textcolor{black}{are} shown in the Table~\ref{dse}, with a power of 0.80 W and a performance of 2.30. Compared with \textcolor{black}{the} engineer's selection, which is also close to the power constraint, it achieves 0.22 higher performance. It demonstrates PANDA's potential in DSE applications. Compared with existing DSE works \textcolor{black}{that} invoke VLSI design flow iteratively, once PANDA is trained with a few known samples, it requires no more sampling or update for the space exploration task.



\subsection{Resource Function and Power Analysis}
\label{sec:resource_ana}

\begin{figure}[!t]
\centering
\vspace{-.15in}
\hspace{-6mm}
\subfigure[D-Cache]{
    \centering
    \includegraphics[height=0.14\textwidth]{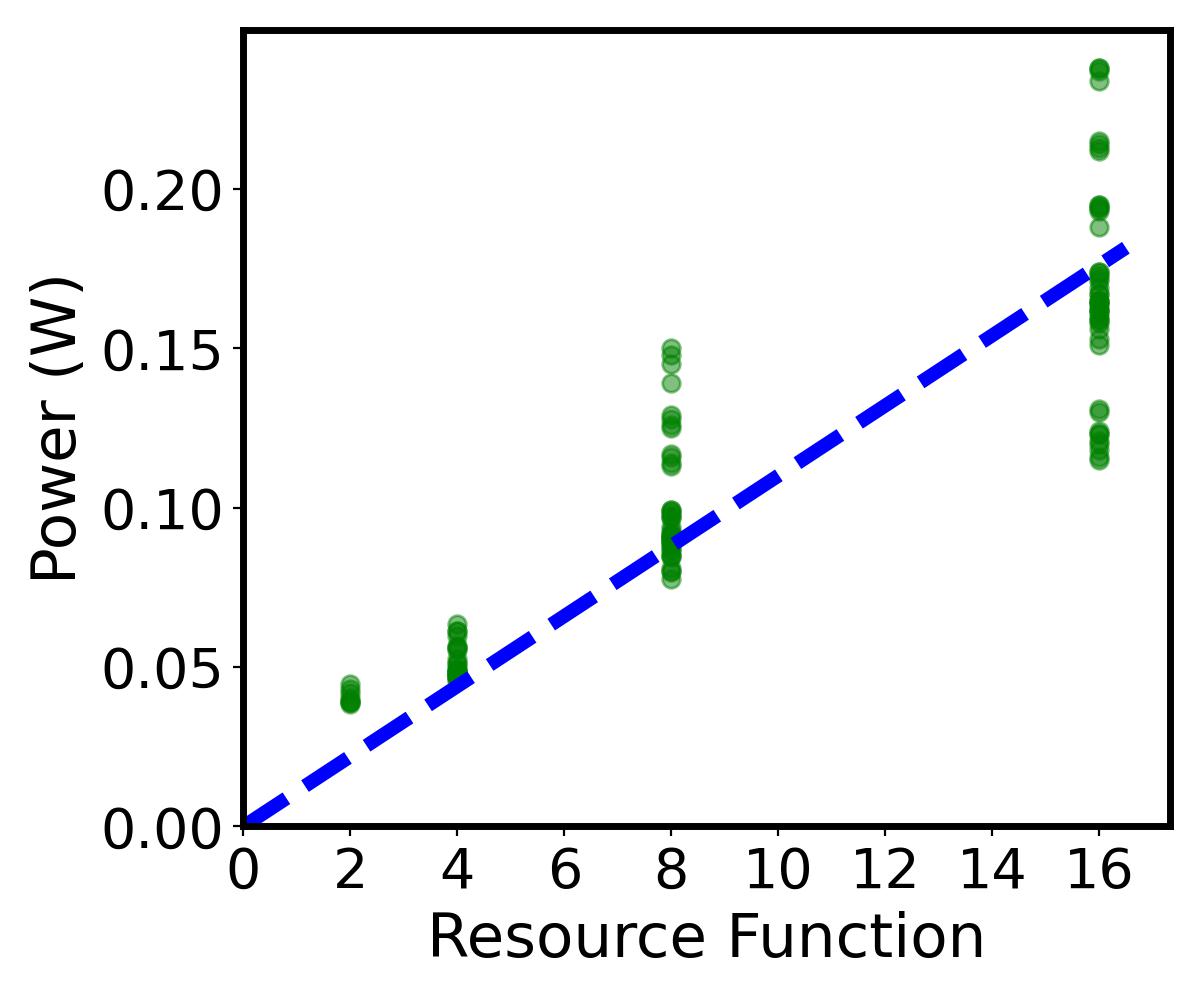}
    \label{dcache_stats}
}
\hspace{-5mm}
\subfigure[ISU]{
    \centering
    \includegraphics[height=0.14\textwidth]{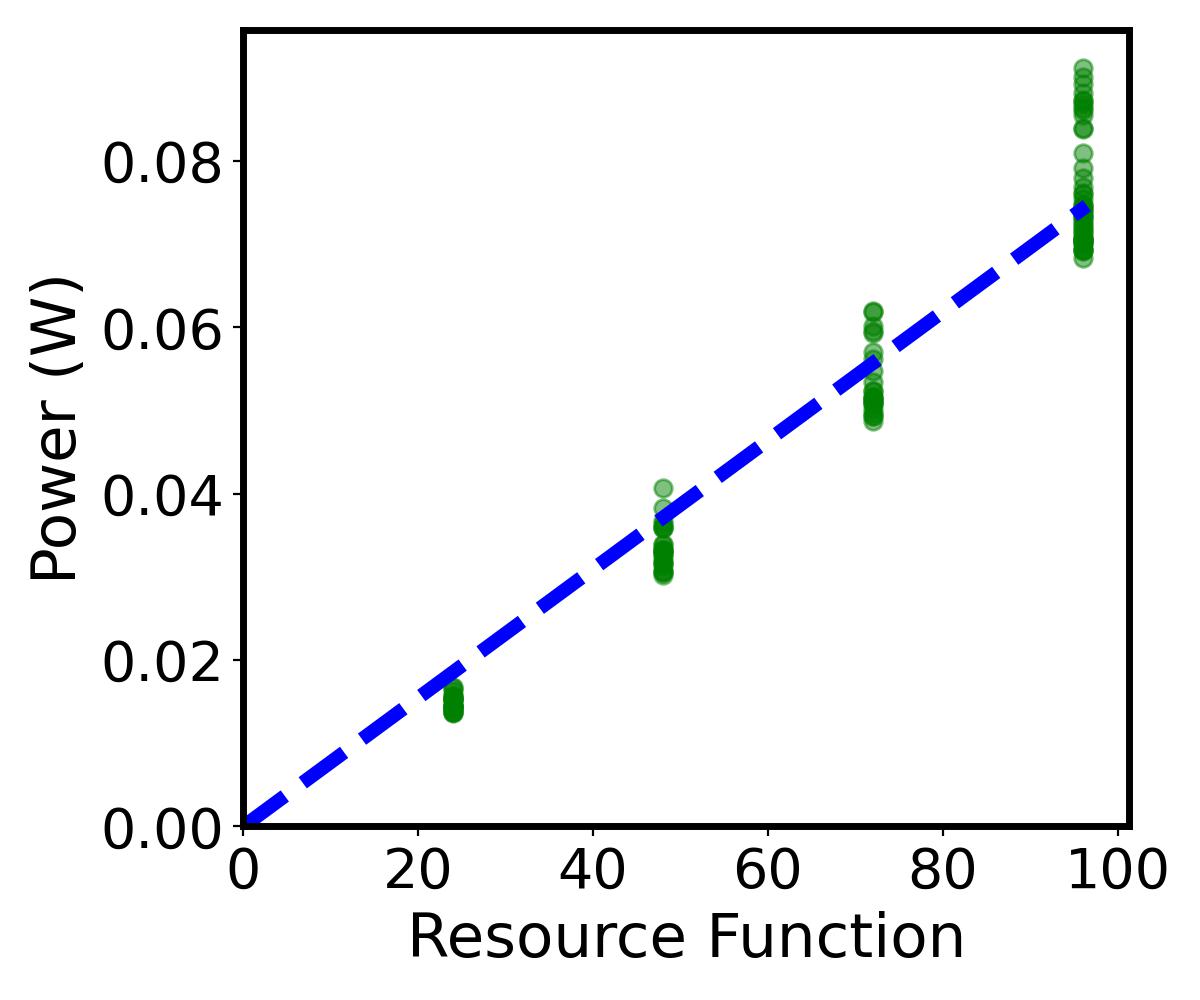}
    \label{isu_stats}
}
\hspace{-5mm}
\subfigure[Other Logic]{
    \centering
    \includegraphics[height=0.14\textwidth]{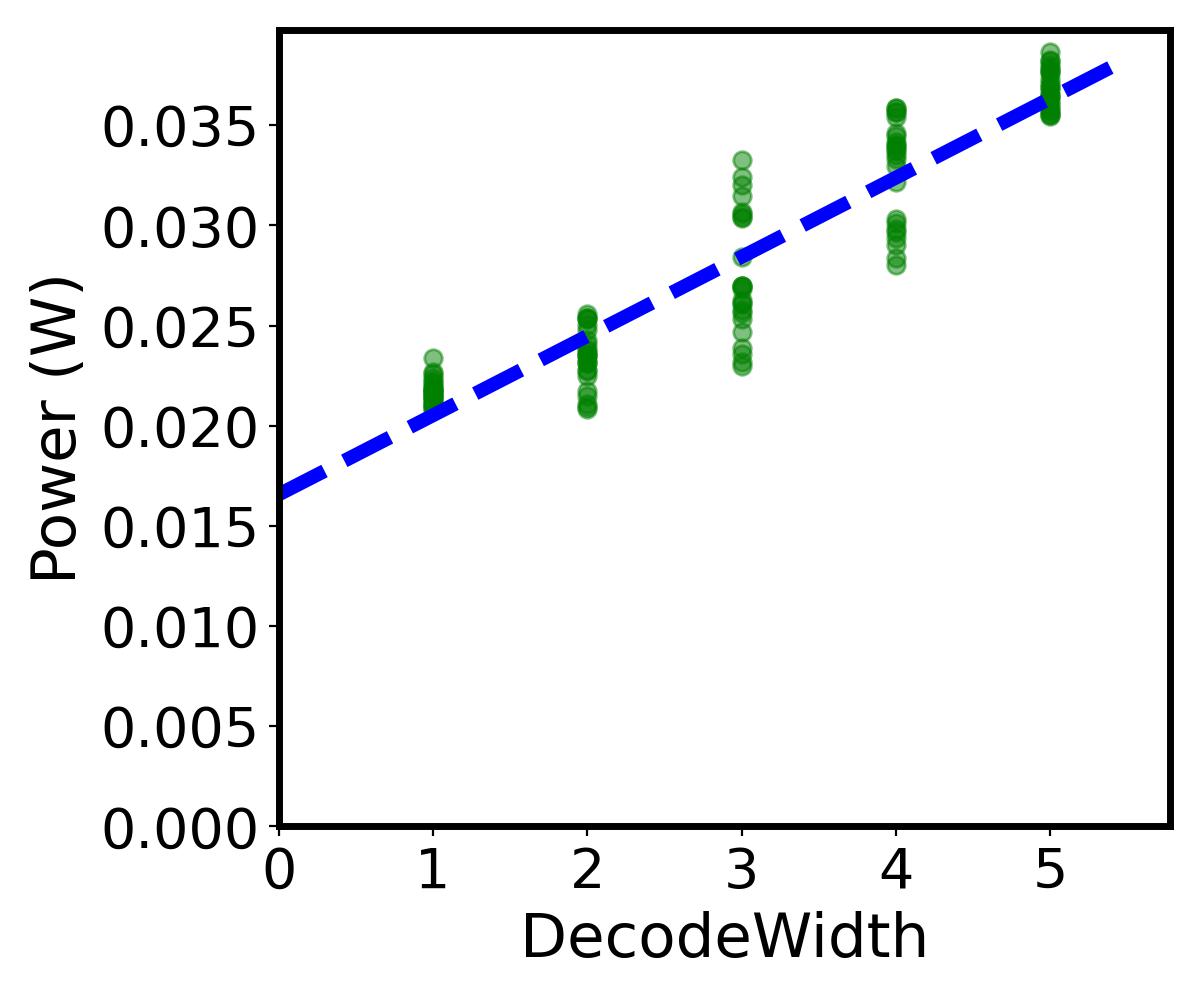}
    \label{logic_stats}
}
\vspace{-.1in}
\caption{Component power vs its resource function $\boldsymbol{F_{res}^i}$, except (c), which demonstrates the relationship between power and DecodeWidth, because the bias in its resource function needs to be estimated.  }
\label{component_result}
\vspace{-.15in}
\end{figure}

Finally, to verify the correctness of PANDA, we first visualize the correlation between ground-truth component power and the corresponding resource function $\boldsymbol{F_{res}}^i$ in Fig.~\ref{component_result}. For D-Cache and ISU, power mainly scales proportionally with the resource function (x-axis). As for the Other Logic, we show \emph{DecodeWidth} in x-axis, and its power is also proportional to the resource function \emph{DecodeWidth} + bias. These data patterns indicate the correctness of PANDA's resource function $\boldsymbol{F_{res}}^i$, which will be multiplied by ML model to generate power prediction, as defined by Equation~(\ref{eq:panda}).


From another perspective, Fig.~\ref{component_result} also shows that when the resource function is fixed in x-axis, there are still many power value variations in \textcolor{black}{the} vertical direction, caused by the difference in other configuration parameters and event parameters. These variations will be captured by PANDA's ML part of each component $\boldsymbol{F_{ml}^i}$. According to Equation~(\ref{eq:panda}), the ML model $\boldsymbol{F_{ml}^i}$ is actually trained to predict the power divided by the resource function (i.e., power$/\boldsymbol{F_{res}}^i$).

\begin{figure}[!t]
\centering
\vspace{-.1in}
\hspace{-6mm}
\subfigure[Original Distribution]{
    \centering
    \includegraphics[height=0.2\textwidth]{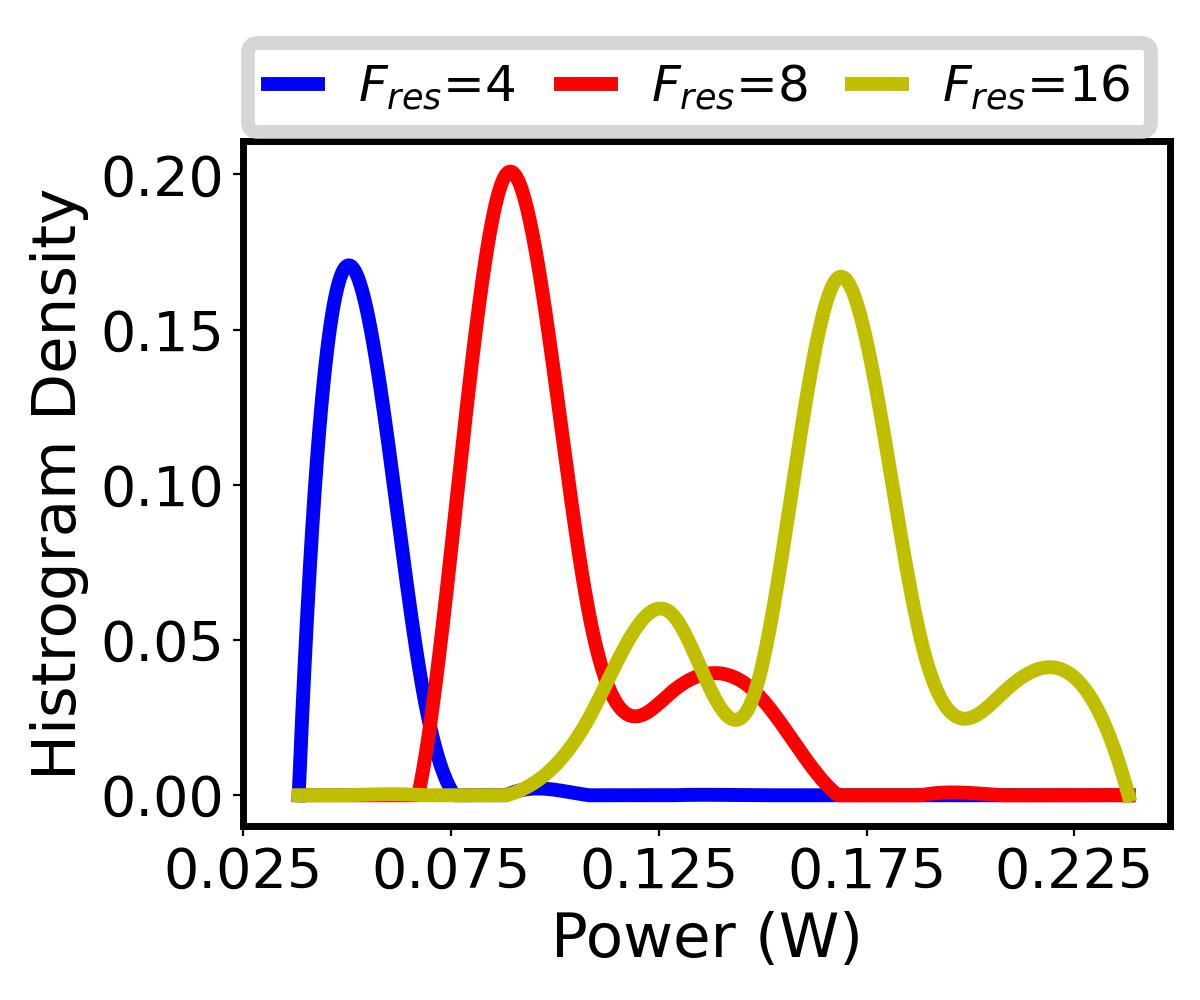}
    \label{dcache_original}
}
\hspace{-5mm}
\subfigure[Distribution for $\boldsymbol{F_{ml}^i}$ to learn]{
    \centering
    \includegraphics[height=0.2\textwidth]{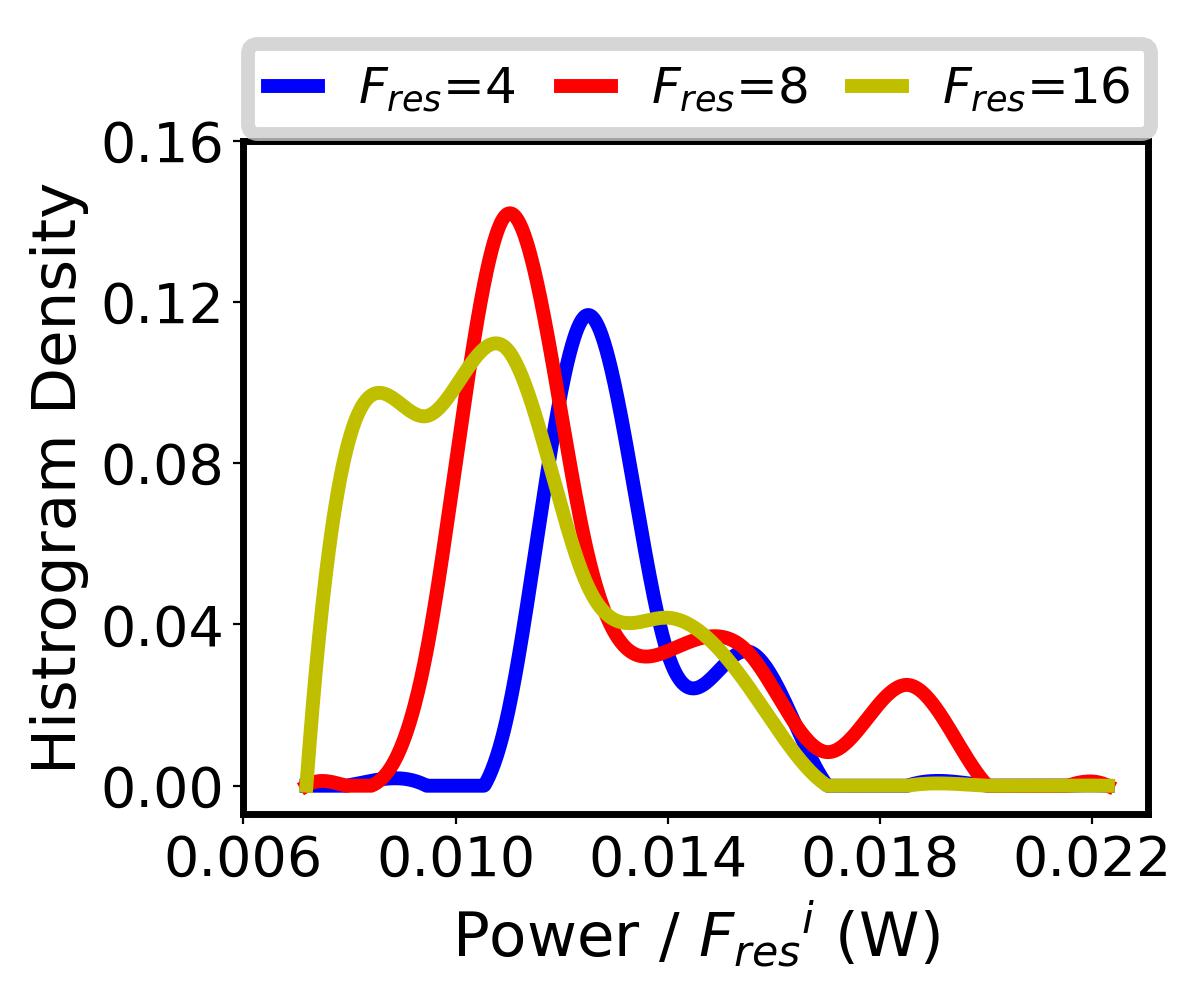}
    \label{dcache_ml}
}
\vspace{-.1in}
\caption{Distribution of D-Cache power, it corresponds to the power distribution of points of Fig.~\ref{component_result}(a). (a) Original power, learned by existing ML methods. (b) Power divided by resource function (power$/\boldsymbol{F_{res}}^i$), learned by PANDA's ML model. PANDA's ML part $\boldsymbol{F_{ml}^i}$ learns more similar distributions, benefiting accuracy when training data is limited.}
\label{power_distribution}
\vspace{-.2in}
\end{figure}

To analyze such vertical power variation for different resource function values, we visualize the power distribution of D-Cache in Fig.~\ref{power_distribution}(a). It corresponds to vertical points in Fig.~\ref{component_result}(a)\footnote{Configurations with $F_{res}=2$ only \textcolor{black}{account} for 6\% among all configurations of D-Cache displayed in Fig.~\ref{component_result}(a). Therefore, we discard this small part in Fig.~\ref{power_distribution}, only showing $F_{res}=4, 8, 16$.}. We further visualize the distribution of power$/\boldsymbol{F_{res}}^i$ in Fig.~\ref{power_distribution}(b). The comparison between Fig.~\ref{power_distribution}(a) and Fig.~\ref{power_distribution}(b) shows an interesting pattern and \textcolor{black}{provides} another explanation of the superior performance of PANDA.

In Fig.~\ref{power_distribution}(a), the power distributions of configurations with different resource function values are largely different. As a result, when training data is limited, ML models may only see training samples from a few distributions, then perform bad on testing designs from unknown other distributions. The gap between different distributions is large, causing an obvious prediction error. In comparison, PANDA actually trains the ML model to predict power$/\boldsymbol{F_{res}}^i$, as shown in Fig.~\ref{power_distribution}(b). This power$/\boldsymbol{F_{res}}^i$ objective provides obviously more similar distributions. Even when training data is limited, the ML part's prediction will fall into a similar distribution anyway, without causing a large error. This analysis \textcolor{black}{provides} one more rationale \textcolor{black}{for} our multiplying ML model with resource function in PANDA.

%% file: _txt/6_conclusion.tex

\section{Conclusion}

In this work, we propose PANDA, an architecture-level power model that unifies analytical and machine learning techniques. PANDA develops its own simpler analytical function for each component based on architecture knowledge,
leaving more complex patterns to be learned by the ML part. It significantly outperforms state-of-the-art solutions, and maintains a high accuracy even with \textcolor{black}{the} very limited number of known configurations for training. Such a data-friendly solution lowers the barrier \textcolor{black}{to} adopting ML techniques by design teams, and thus PANDA is a compelling addition to the \textcolor{black}{architects’} toolbox.

\section{Acknowledgement}

This work is partially supported by the Hong Kong Research Grants Council (RGC) ECS Grant 26208723, Semiconductor Research Corporation (SRC) GRC-CADT 3103.001/3104.001, and ACCESS – AI Chip Center for Emerging Smart Systems, sponsored by InnoHK funding, Hong Kong SAR. Also, the authors thank the help
from Prof Bei Yu and Chen Bai at the Chinese University of Hong Kong.

